\documentclass[final]{cvpr}

\usepackage{times}
\usepackage{epsfig}
\usepackage{graphicx}
\usepackage{amsmath}
\usepackage{amssymb}
\usepackage{appendix}

\usepackage{multirow}

\usepackage[pagebackref=true,breaklinks=true,colorlinks,bookmarks=false]{hyperref}

\def\cvprPaperID{****} 
\def\confYear{CVPR 2021}

\usepackage{booktabs} 
\usepackage{bbding}

\usepackage{pifont}
\usepackage{wasysym}
\usepackage{amssymb}
\usepackage[bottom]{footmisc}

\begin{document}

\title{NAPA: Neural Art Human Pose Amplifier} 

\author{Qingfu Wan\\
New York University\\
{\tt\small qw894@nyu.edu}
\and
Oliver Lu\\
New York University\\
{\tt\small ozl201@nyu.edu}
}

\maketitle

\begin{abstract}
This is the project report for \href{https://cs.nyu.edu/~fergus/teaching/vision/index.html}{CSCI-GA.2271-001}. We target human pose estimation in artistic images. For this goal, we design an end-to-end system that uses neural style transfer for pose regression. We collect a 277-style set for arbitrary style transfer and build an artistic 281-image test set. We directly run pose regression on the test set and show promising results. For pose regression, we propose a 2d-induced bone map from which pose is lifted. To help such a lifting, we additionally annotate the pseudo 3d labels of the full in-the-wild MPII dataset. Further, we append another style transfer as self supervision to improve 2d. We perform extensive ablation studies to analyze the introduced features. We also compare end-to-end with per-style training and allude to the tradeoff between style transfer and pose regression. Lastly, we generalize our model to the real-world human dataset and show its potentiality as a generic pose model. We explain the theoretical foundation in \textcolor{magenta}{Appendix}. We release code at \url{https://github.com/strawberryfg/NAPA-NST-HPE}, ~\href{https://drive.google.com/drive/folders/1omDWZeG6zA8GJx5Ij9Y1qJZiY8YYTcFx?usp=sharing}{data}, and  ~\href{https://drive.google.com/file/d/1C0pw2HUWV5jHSRNZ-zBP8hW7f8Dl5HZ7/view?usp=sharing}{video}.

\end{abstract}

\section{Introduction}

The aim of human pose estimation is to locate pre-defined keypoints of human figures. Prior work has focused on real-world datasets. There is limited work addressing 3d pose estimation in artistic images.


%

Neural style transfer can effectively transfer the style of phenomenal artworks onto real images. However, to emulate the different aesthetic effects of artworks \eg \emph{paintings,  sculptures}, existing style sets are not enough. Regarding our application, we build a style set covering different styles and enforce novel style transfer losses. We then empower our pose system with style transfer and attach an auxiliary style transfer in the end to further improve 2d keypoints. 

To enable fine pose regression, we decouple the problem into 2d and depth using bone maps. To facilitate the learning of depth therein, we annotate pseudo 3d labels of MPII.

We are the first to showcase high-quality 2d/3d pose predictions on artwork about human figures.

We make additional contributions:

\begin{itemize}
    \item [$\bullet$] We investigate end-to-end $vs$ per-style training and compare stylization results. We identify the importance of instance normalization. We display more impressive stylized images using the per-style training strategy and establish a solid baseline.
    
    \item [$\bullet$] We show that our method can be extended to generic pose estimation regardless of genuineness. \footnote{\emph{"As Apollinairre remarked, a chair will be understood as a chair from no matter what point of view it is seen if it has the essential components of a chair."}\cite{fry1978cubism} A human figure, whether real or artistic, has the elements to be recognized by vision algorithms.}
    
    \item[$\bullet$] We perform comprehensive ablation studies to expound on all of the introduced features/components.
    
    \item[$\bullet$] We provide theoretical analysis in \textcolor{magenta}{Appendix}.
\end{itemize}

\section{Related Works}

\textbf{Style Transfer} \cite{gatys2016image} introduced neural style transfer by applying convolutional neural nets to reproduce famous painting styles on natural images. The algorithm combined the content of a natural image with the style of a famous painting by minimizing both the feature and style reconstruction losses. The algorithm produced high-quality images, but was computationally expensive as it required both forward and backward passes through a pre-trained network. \cite{johnson2016perceptual} developed a feed-forward method of neural style transfer that is much faster than the method introduced by \cite{gatys2016image} with similar qualitative results. Our method for style transfer in this paper is based on \cite{johnson2016perceptual}.

\textbf{Human Pose Estimation} Early works established mathematical models \eg exponential maps \cite{bregler1998tracking}, shape context\cite{mori2002estimating}, Pfinder \cite{wren1997pfinder}, deformable part models\cite{felzenszwalb2008discriminatively}, \etc. Deep learning has completely revolutionized this field\cite{li2020cascaded}\cite{kundu2020self}\cite{habibie2019wild}\cite{omran2018neural}, we refer to \cite{chen2020monocular} for an up-to-date survey. \cite{jenicek2019linking} used pose in art composition for image retrieval. \cite{madhu2020understanding} analyzed artwork compositions via pose. The closest to our work is a very recent work \cite{madhu2020enhancing} that used style transfer for pose estimation/image retrieval in Greek vase paintings.

\section{Approach}
\begin{figure*}
\begin{center}
   \includegraphics[width=1.0\linewidth]{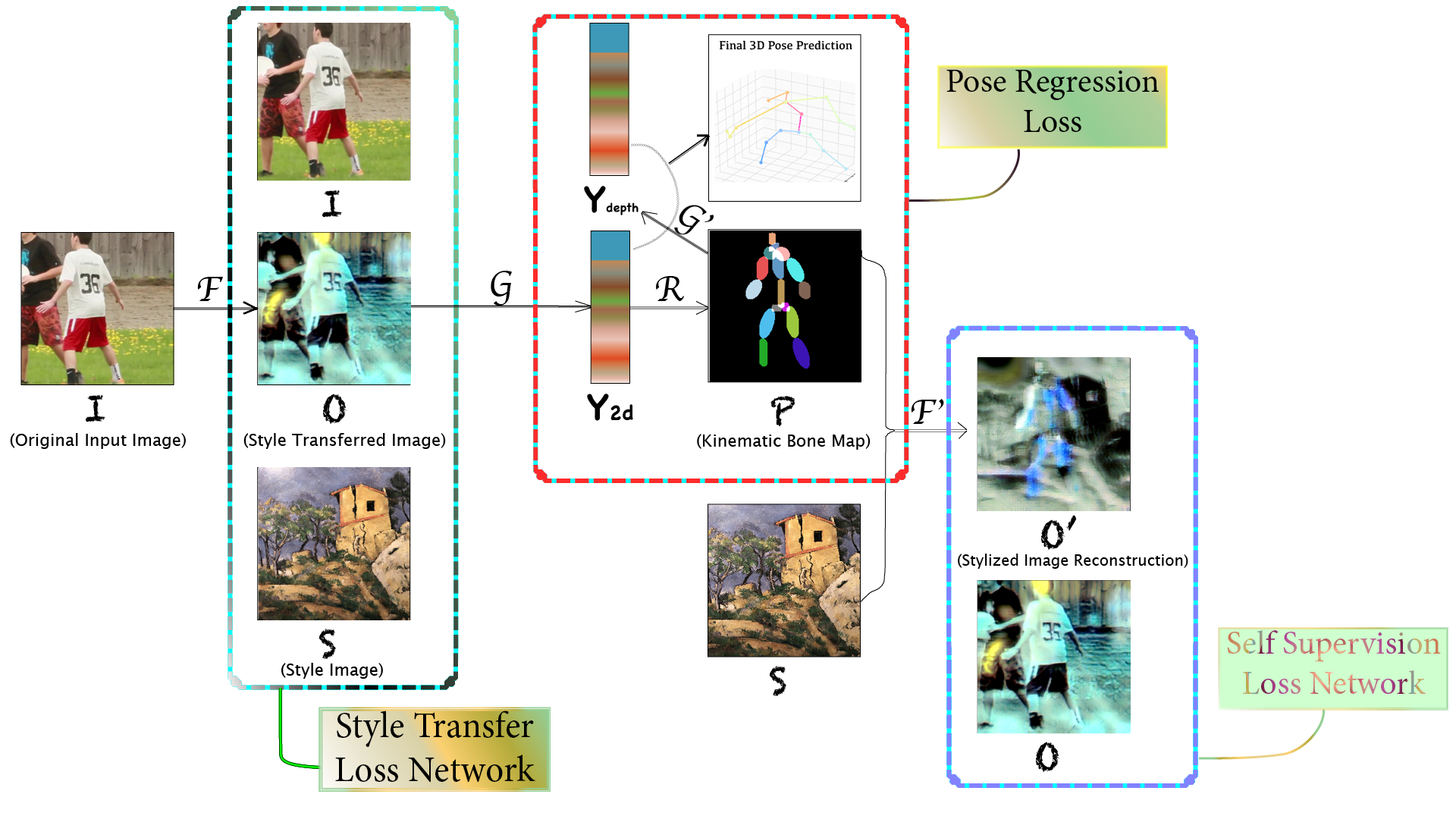}
   
\end{center}
   \caption{Diagram of the pipeline. During training, the neural style transfer (\textcolor{green}{green}) part stylizes the human image and sends the stylization to the pose regression part (\textcolor{red}{red}), which outputs the final 3d pose prediction as desired. The self supervision part (\textcolor{blue}{blue}) aims to improve only the 2d keypoints and can be treated as additional supervision. During testing, the original art image is directly sent to the pose regression part which predicts the 3d pose. The style transfer and self supervision parts are completely discarded.}
   \label{fig:pipeline}
\end{figure*}

We sketch the overall pipeline in Fig. ~\ref{fig:pipeline}. It consists of three parts: \textbf{the neural style transfer part} (\textcolor{green}{green} in Fig. ~\ref{fig:pipeline}), \textbf{the pose regression part} (\textcolor{red}{red}), and \textbf{the self supervision part} (\textcolor{blue}{blue}). 
During training, the original image containing a person $\mathbf{I}$ is presented to the image transformation network $\mathcal{F}$, which then produces a style transferred image $\mathbf{O}$. The following pose regression network $\mathcal{G}$ takes only the stylized image $\mathbf{O}$ as input and regresses the 3d keypoints $\mathbf{Y}$. Note this is the final output we want: the 3d pose prediction. The extra self supervision part reuses the concept of neural style transfer and tries to reconstruct the stylization $\mathbf{O}$ from the style image $\mathbf{S}$ and the bone map $\mathbf{P}$ (which is derived from 2d keypoints $\mathbf{Y}_{2d}$).
The test stage merely maps $\mathbf{I}$ to 3d pose $\mathbf{Y}$ using $\mathbf{G}$. The notation overview is in \textcolor{magenta}{Appendix Sec. A}.

\textbf{Style Transfer} This process stylizes the content image to emulate the artistic effect of the style target. For this stylization, we use the same image transform net $\mathcal{F}$ as \cite{johnson2016perceptual}. Regarding the style image $\mathbf{S}$, we randomly select from a pool we collected (Sec. ~\ref{sec:artstyle}) instead of training one model per style as in \cite{johnson2016perceptual}. 
We also tried the approach in \cite{johnson2016perceptual} and detail the differences between \emph{style-specific training} and \emph{arbitrary style training} below in Sec. ~\ref{sec:e2etrain}. Following \cite{johnson2016perceptual} we utilize a VGG as the loss network and enforce more losses to guide the style transfer process. (Sec. ~\ref{sec:lossfunc}, ~\ref{sec:losses}). 

\textbf{Pose Regression} After transferring, we would like to generate the 3d pose (also called \emph{3d keypoints}) $\mathbf{Y} \in \mathbb{R}^{\mathit{J} \times 3}$ ($\mathit{J}=18$) of that human figure. The problem is decoupled into estimating 2d and lifting 2d to 3d. We extract 2d keypoints $\mathbf{Y}_{2d}$ from $\mathcal{G}$: \cite{sun2018integral} here. Similar to \cite{wan2017deepskeleton} , we visualize the 2d in a kinematic bone map $\mathbf{P}$ and forward it to another depth estimation network $\mathcal{G}^\prime$: (a simple Res50 with 1 FC regression head) to obtain depth $\mathbf{Y}_{depth}$. Now we can piece together $\mathbf{Y}_{2d}$ with $\mathbf{Y}_{depth}$ to construct the 3d pose $\mathbf{Y}$. We empirically find this decoupling and the proposed bone map benefit test cases which would otherwise cause bizarre results. (Fig. ~\ref{fig:figske}, Sec. ~\ref{sec:ske}) Losses are in Sec. ~\ref{sec:lossfunc}
.

The bone map rendering function $\mathcal{R}$ runs as follows: For each bone in the kinematics chain \cite{zhou2016deep} connecting $2$ adjacent keypoints $(x_1,y_1)$ and $(x_2,y_2)$, we define the related area in the bone map as the oval (\emph{and its interior}) that has $\overrightarrow{(x_2,y_2)-(x_1,y_1)}$ as the minor axis $\vec{y}$, the major axis   $\vec{x} \perp \vec{y}$ and bone width $d$ as the length of the major axis. $d$ effectively controls how wide we want the bone to span. The oval contains a set of loops $[\alpha]$: $\lbrace \hat{\alpha_i}: [0,1] \rightarrow \mathit{R}^2 \mid \hat{\alpha_i}\ is\ the \ oval\ \frac{(dist((x,y), \vec{y}))^2 }{a_i^2} + \frac{(dist((x,y),\vec{x}))^2}{(\frac{dist((x_1,y_1),(x_2,y_2))}{2})^2} = 1,\ for\ 0\leq a_1<a_2<...<a_i<...<a_n=\frac{d}{2} \rbrace$ A more formalized representation of the time-parameterized mapping between $[0,1]$ to the bone map space $\mathcal{P}$ is detailed in \textcolor{magenta}{Appendix Sec. K}. The bone map is finalized simply by colorizing each bone area illustrated above in a 3-channel RGB. The rendering can be better understood from the angle of loops in fundamental groups. (\textcolor{magenta}{Appendix Sec. I.2})

\textbf{Self Supervision} The appended self supervision loss network corrects the predicted 2d pose by projecting it to the stylization space $\mathcal{O}$ and measuring the difference between the reconstruction $\mathbf{O}^\prime \in \mathcal{O}$ and the stylized image $\mathbf{O} \in \mathcal{O}$. We adopt another image transform net $\mathcal{F}^\prime$, which is initialized to $\mathcal{F}$, to reconstruct the stylization given the bone map $\mathbf{P}$ above and the style target $\mathbf{S}$. Essentially it can be interpreted as another style transfer on the rendered bone map.  
We explain more details and the reason why we do not reconstruct the original image in \textcolor{magenta}{Appendix. Sec. I}.

\subsection{Artistic Styles} 
\label{sec:artstyle}
Our pipeline supports arbitrary style transfer, which is crucial for separating pose from nuisances. Previous style transfer works only have a few most recognized paintings from most well-known painters \footnote{"The Starry Night" by Vincent Van Gogh, "Woman with a Hat" by Henri Matisse, 
"Ma Jolie
" by Pablo Picasso \etc}, which is limited. We build a style dataset of \textbf{$277$} images covering different styles \footnote{\url{https://drive.google.com/drive/folders/1zNma_7vXZa_lwHOQtRrSb7NIiGw_38Xc?usp=sharing}}, which we collect from museums, galleries and books \footnote{~\href{https://www.moma.org/}{MoMA}, ~\href{https://www.sfmoma.org/stories/}{SFMOMA}, ~\href{https://whitney.org/}{Whitney Museum}, ~\href{https://www.guggenheim.org/}{Guggenheim Museum},
~\href{http://www.mocashanghai.org/?lang=en}{MoCA Shanghai}, ~\href{http://en.chnmuseum.cn/}{NMC Beijing} \etc; ~\href{https://www.lissongallery.com/}{Lisson Gallery}, ~\href{https://www.kasmingallery.com/}{Kasmin Gallery} \etc; \href{https://www.goodreads.com/book/show/15893647-steampunk}{"steampunk the beginning"}, ~\href{https://www.goodreads.com/book/show/374507.Masterpieces_of_Impressionism_and_Post_Impressionism}{"Masterpieces of impressionism \& post-impressionism"}, ~\href{https://www.goodreads.com/book/show/5505587-expressionism}{"Expressionism"}, ~\href{https://www.goodreads.com/book/show/2756643-gothic-art}{"Gothic art"}, ~\href{https://www.goodreads.com/book/show/8273444-cubism}{"Cubism"} \etc}.  We use styles ranging from Impressionism, Cubism, Fauvism, to  Romanesque, Futurism, Surrealism, Expressionism, Kinetic Art \etc. Some examples are in \textcolor{magenta}{Appendix Sec. C}. We also show some failure cases under \href{https://github.com/strawberryfg/NAPA-NST-HPE/tree/main/train/per-style-training}{this page}.

\subsection{Losses}
\label{sec:lossfunc}
\textbf{Neural Style Transfer Loss}



\begin{itemize}

\item[$\bullet$]

$\mathcal{L}_{style}(\mathrm{O},\mathrm{S}) $: It is the squared Frobenius norm of the difference between the Gram matrices of the style transferred image and the style image \cite{johnson2016perceptual}.

\item[$\bullet$]

$\mathcal{L}_{feat}(\mathrm{O}, \mathrm{I}) $: The feature reconstruction loss between content and stylized image in \cite{johnson2016perceptual}.

\item[$\bullet$]

$\mathcal{L}_{tv}(\mathrm{O}) $: Total variation loss as in \cite{johnson2016perceptual}.
\end{itemize}

Note the above are used in the perceptual loss \cite{johnson2016perceptual} paper.

Besides, we propose the following:

\begin{itemize}

\item[$\bullet$]

$\mathcal{L}_{sent}(\mathrm{O},\mathrm{S}) $: Jensen Shannon entropy loss between features from stylized image $\mathrm{O}$ and style image $\mathrm{S}$ at layers $\mathbf{relu1\_1}$,
$\mathbf{relu1\_2}$ and
$\mathbf{relu2\_1}$.

\item[$\bullet$]

$\mathcal{L}_{srgb}(\mathrm{O_{srgb}}, \mathrm{S_{srgb}})  $: Let $\mathrm{O_{srgb}},\mathrm{S_{srgb}} $ be the srgb format of transferred image and style image respectively, the loss is the cosine similarity of these two vectors $\mathrm{O_{srgb}}$ and $\mathrm{S_{srgb}}$. Note losses on other color spaces \cite{DBLP:journals/corr/abs-1902-00267}   might further help, we leave that for future work.

\item[$\bullet$]

$\mathcal{L}_{hsv}(\mathrm{O}, \mathrm{S}) $: This is the HSV loss between style transferred image $\mathrm{O}$ and style image $\mathrm{S}$. Let the hue, saturation and vibrance of style image be $\mathrm{h_s}, \mathrm{s_s},\mathrm{v_s}$ and the counterparts of the transferred image be $\mathrm{h_o}, \mathrm{s_o},\mathrm{v_o}$. $\mathcal{L}_{hsv} = || \mathrm{h_s} - \mathrm{h_o}||_1 + || \mathrm{s_s} - \mathrm{s_o}||_1 + || \mathrm{v_s} - \mathrm{v_o}||_1 $.

\item[$\bullet$]

$\mathcal{L}_{cos} (\mathrm{O},\mathrm{S})$: Defined as the cosine similarity between features from style image $\mathrm{S}$ and transferred image $\mathrm{O}$ at layer $\mathbf{relu4\_2}$.
\end{itemize}


\begin{itemize}
\item[$\bullet$] $\mathcal{L}_{content}(\mathrm{O},\mathrm{I})$: The L2 loss between features from content image $\mathrm{I}$ and stylized image $\mathrm{O}$ at layer $\mathbf{relu4\_2}$.

\item[$\bullet$]

$\mathcal{L}_{cent}(\mathrm{O},\mathrm{I})  $:
Written as the Jensen Shannon entropy loss between features from content image and stylized image at layers $\mathbf{relu5\_2}$, 
$\mathbf{relu5\_3}$,
$\mathbf{relu5\_4}$

\end{itemize}

To summarize, $\mathcal{L}_{sent} $, $\mathcal{L}_{srgb}$, $\mathcal{L}_{hsv}$ and $\mathcal{L}_{cos}$ relate to \emph{style reconstruction loss} while $\mathcal{L}_{content}$ and $\mathcal{L}_{cent}$ represent the \emph{content reconstruction loss}. Practically we observe that among the additional losses apart from those in \cite{johnson2016perceptual}, the following are the most important: $\mathcal{L}_{hsv}$, $\mathcal{L}_{cos}$ and $\mathcal{L}_{srgb}$. We will see this  afterwards in Sec. ~\ref{sec:losses}.


\textbf{Pose Regression Loss}
As said earlier, we subscribe to the view that lifting from 2d to 3d is better than directly regressing 3d when a generic pose model is required (Sec. ~\ref{sec:crossdataset}), and so the losses are on 2d and depth respectively.

$\mathcal{L}_{2d}$: 
Integral loss guides the learning of $\mathcal{G}$ and 2d keypoints prediction $\mathbf{Y}_{2d}$(Fig. ~\ref{fig:pipeline}). It was first put forth in \cite{sun2018integral}.

$\mathcal{L}_{depth}$: We use the orientation representation\cite{luo2018orinet}\cite{sun2017compositional}\cite{wan2019patch} and define an euclidean loss  between predicted bone vector and ground truth bone vector. This loss forces the model to implicitly learn the depth of keypoints $\mathbf{Y}_{depth}$ by transforming bone vector to joint location. Bone vector is standardized (using \emph{mean} and \emph{std}) to facilitate the learning.

\textbf{Self Supervision Loss} The self supervision is a small "gadget" to refine the 2d by comparing the induced bone map and the stylized image in the stylization space. To initiate the comparison, we adapt the ones in \emph{neural style transfer loss} above and define the counterparts:

\begin{itemize}
\item[$\bullet$]
$\mathcal{L}_{style\_sup}(\mathrm{O'},\mathrm{O}) $: Similar to $\mathcal{L}_{style}$ except that the difference is between the stylized image: $\mathrm{O}$ and the reconstruction of the stylized image: $\mathrm{O'}$.

\item[$\bullet$]
$\mathcal{L}_{cos\_sup}(\mathrm{O'}, \mathrm{O})$: The cosine similarity between $\mathrm{O}$ and its reconstruction $\mathrm{O'}$. It is similar to $\mathcal{L}_{cos}$.

\item[$\bullet$]
$\mathcal{L}_{feat\_sup}(\mathrm{O'}, \mathrm{O})$: Now the \emph{feature} to be reconstructed is $\mathrm{O}$. It is defined at the content layer $\mathbf{relu4\_2}$.
\end{itemize}

\textbf{Ratios} The learning is all about balancing, most importantly the balancing between the neural style transfer and the pose regression. This is critical for the end-to-end training mechanism in our pipeline, which is not needed for a separate training alternative (Sec. ~\ref{sec:e2etrain}). We empirically set the ratios during training. See \textcolor{magenta}{Appendix Sec. B}.

\subsection{Training}
\label{sec:training}
The entire pipeline includes a lot of prevalent network architectures. Direct end-to-end training from scratch does not work as there is a slew of losses to balance. In our exploration, we follow a 4-stage training protocol:

\emph{stage 1} only learns the style transfer network $\mathcal{F}$ with VGG loss network tunable.

\emph{stage 2} only targets at regressing the pose. We warm up the $\mathcal{G}$(\emph{2d}) with \cite{sun2018integral}, and $\mathcal{G'}$(\emph{depth}) with ResNet50. Images are stylized using weights learned from \emph{stage 1}.

\emph{stage 3} fixes the VGG weight and jointly trains \emph{style transfer} ($\mathcal{F}$) and \emph{pose regression} ($\mathcal{G}$ \& $\mathcal{G'}$).

\emph{stage 4} adds self supervision loss training. Another style transfer network $\mathcal{F^\prime}$ and another VGG are trained. 

We use Human3.6M\cite{h36m_pami}, MPII\cite{andriluka20142d} in training, same as \cite{sun2017compositional}\cite{sun2018integral}.
Concerning the pose regression part, we additionally annotate the pseudo 3d ground truth of the entire MPII \footnote{\url{https://github.com/strawberryfg/NAPA-NST-HPE/tree/main/annotation\_tools/mpii\_annotator}} by manually revising the model-optimized 3d while preserving projection \& certain anthropometric constraints \eg joint angle limits\cite{akhter2015pose}.
This way, the information of various \emph{athletic} 3d poses in MPII is retained and best utilized, which is underexplored in prior works. (Sec. \ref{sec:pseudo}, \textcolor{magenta}{Appendix Sec. E}) About the self supervision, we also make use of the vast amount of human images in COCO \cite{lin2014microsoft}(\emph{train2017/val2017}). All images are resized to $224\times 224$. No data augmentation is enabled. Training is mostly done on an 8GB GTX 1070 or an 8GB RTX 2080 with batch size = $2$, $3$ or $22$ depending on the \emph{stage}. Learning rate is initially $0.0001$, which is then decreased upon loss plateau or increment of the stage. We opt for RMSProp with a weight decay of $0.00001$ and momentum set to $0.9$.

\section{Experiments}

\textbf{Evaluation Dataset} We create a dataset of $281$ images that contains artistic style human figures, most of which come from ~\href{https://www.moma.org/}{Museum of Modern Art}, ~\href{https://www.fantasticfiction.com/p/tim-powers/steampunk-beginning.htm}{steampunk: the beginning}, \href{https://www.goodreads.com/book/show/2756643-gothic-art}{
Gothic Art} and \href{https://www.goodreads.com/book/show/374507.Masterpieces_of_Impressionism_and_Post_Impressionism}{Masterpieces of Impressionism and Post Impressionism: The Annenberg Collection}. We develop an \href{https://github.com/strawberryfg/NAPA-NST-HPE/tree/main/annotation_tools/art_img_annotator}{annotation tool} to annotate the 2d labels. Following the standard, we report results using the PCKh metric that measures the percentage of joints for which the prediction is within a certain threshold to the ground truth. The threshold is set to 25\% of the head size throughout the paper. Note this is a strict threshold that requires high-precision estimation. Regarding the 3d pose, since it's infeasible to obtain it without wearable IMUs or marker-based MoCap systems, we only depict qualitative visualization.


\subsection{Baseline}

We perform per-style training using the style transfer network and methodology from \cite{johnson2016perceptual} as a baseline for our neural style transfer to compare against the end-to-end training. For the per-style training, we used the MPII dataset. The images are cropped to focus on the human figures and then resized to $256 \times 256$. Training was done on K80 Tesla with a batch size of 4. We used Adam as the optimizer with a learning rate of 0.001. We train a style transfer network for each style and then the MPII dataset is fed through the trained network to generate a style transferred MPII dataset. This style transferred dataset is then used to train a pose regression model for the specific style. We will discuss the differences between the results from the per-style training and end-to-end training in Sec. ~\ref{sec:e2etrain}.

\subsubsection{Instance Normalization}
\label{sec:instanceNorm}
\begin{figure}
\begin{center}
   \includegraphics[width=1.0\linewidth]{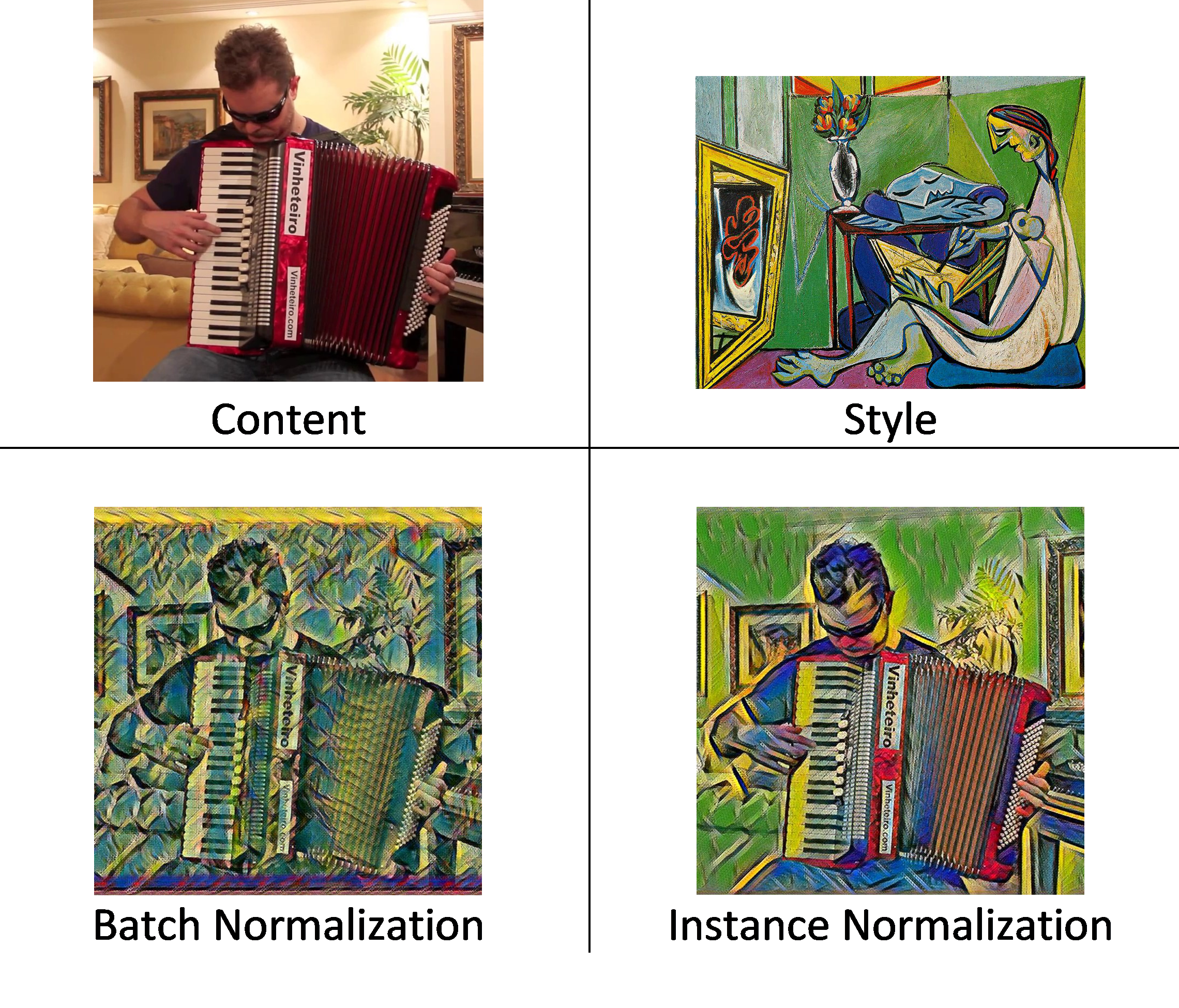}
   
\end{center}
   \caption{Stylized image using batch and instance normalization. Instance normalization improves the production quality.}
   
   \label{fig:instanceNorm}
\end{figure}

The image transformation net from \cite{johnson2016perceptual} that we used has batch normalization layers after the residual convolution layers. Instance normalization has been shown to generate better qualitative images for fast style transfer than batch normalization with improved convergence speeds. \cite{ulyanov2016norm} Instance normalization differs from batch normalization in that every image is normalized separately in contrast to in a batch for batch normalization. As we can see from Fig. ~\ref{fig:instanceNorm}, instance normalization provides for better style transfer than batch normalization as the batch normalization image is mostly the same color and has artifacts around the edges.

\subsection{Visualization}
\begin{figure}[htbp]
\begin{center}
   \includegraphics[width=1.075\linewidth]{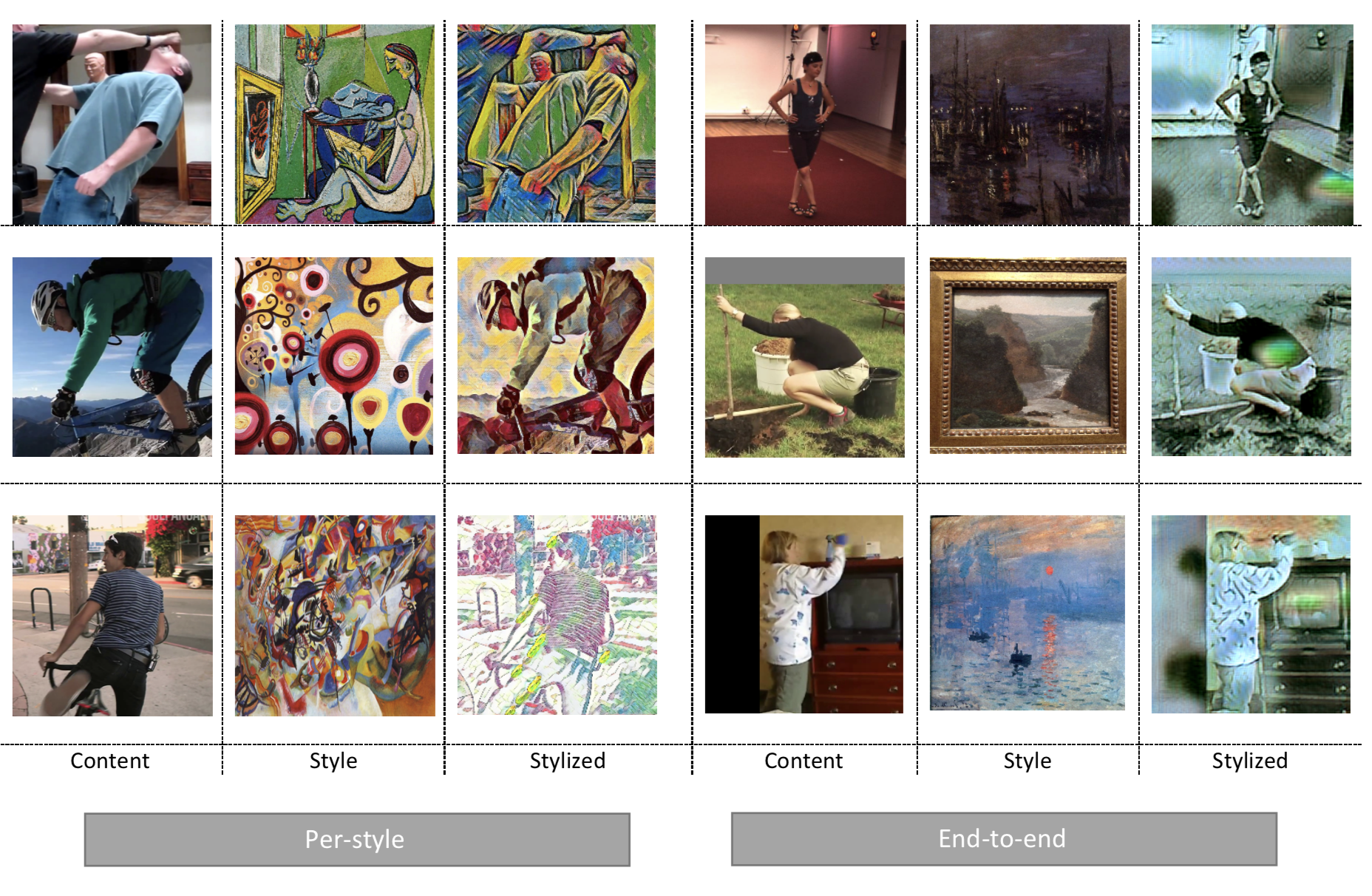}
   
\end{center}
   \caption{Visualization of style transferred images.}
   \label{fig:stylized}
\end{figure}

\begin{figure}
\begin{center}
   \includegraphics[width=0.85375\linewidth]{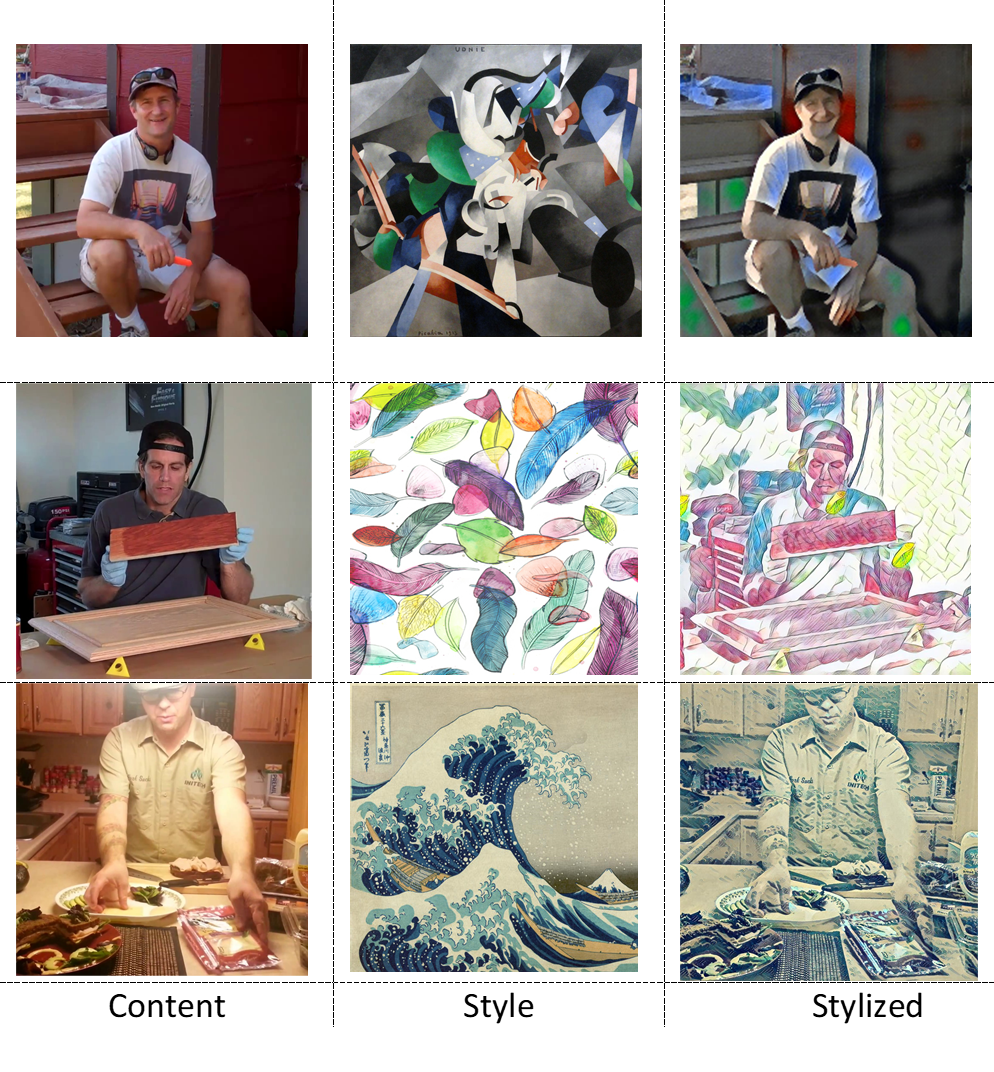}
   
\end{center}
   \caption{Per-style stylization.}
   
   \label{fig:figstyleper2}
\end{figure}

Fig. ~\ref{fig:stylized} and Fig. ~\ref{fig:figstyleper2} demonstrate style transferred images on MPII. More details are in Sec. ~\ref{sec:e2etrain} and here \footnote{\url{https://github.com/strawberryfg/NAPA-NST-HPE/tree/main/train/per-style-training}}. 

\begin{figure}
\begin{center}
   \includegraphics[width=0.95\linewidth]{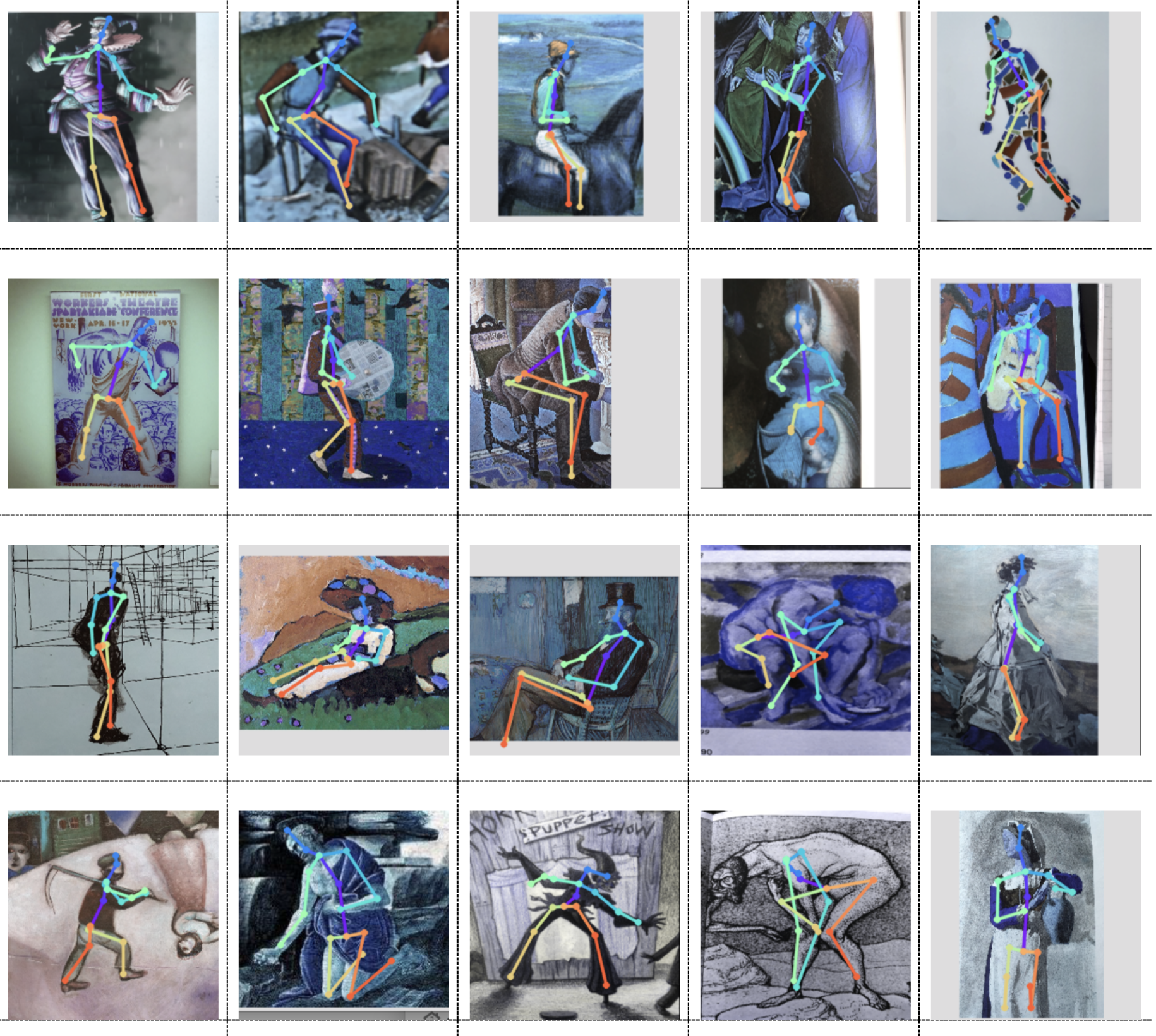}
   
\end{center}
   \caption{2d pose predictions on the test set.}
   \label{fig:posepred2d}
\end{figure}

\begin{figure*}
\begin{center}
   \includegraphics[width=1.0625\linewidth]{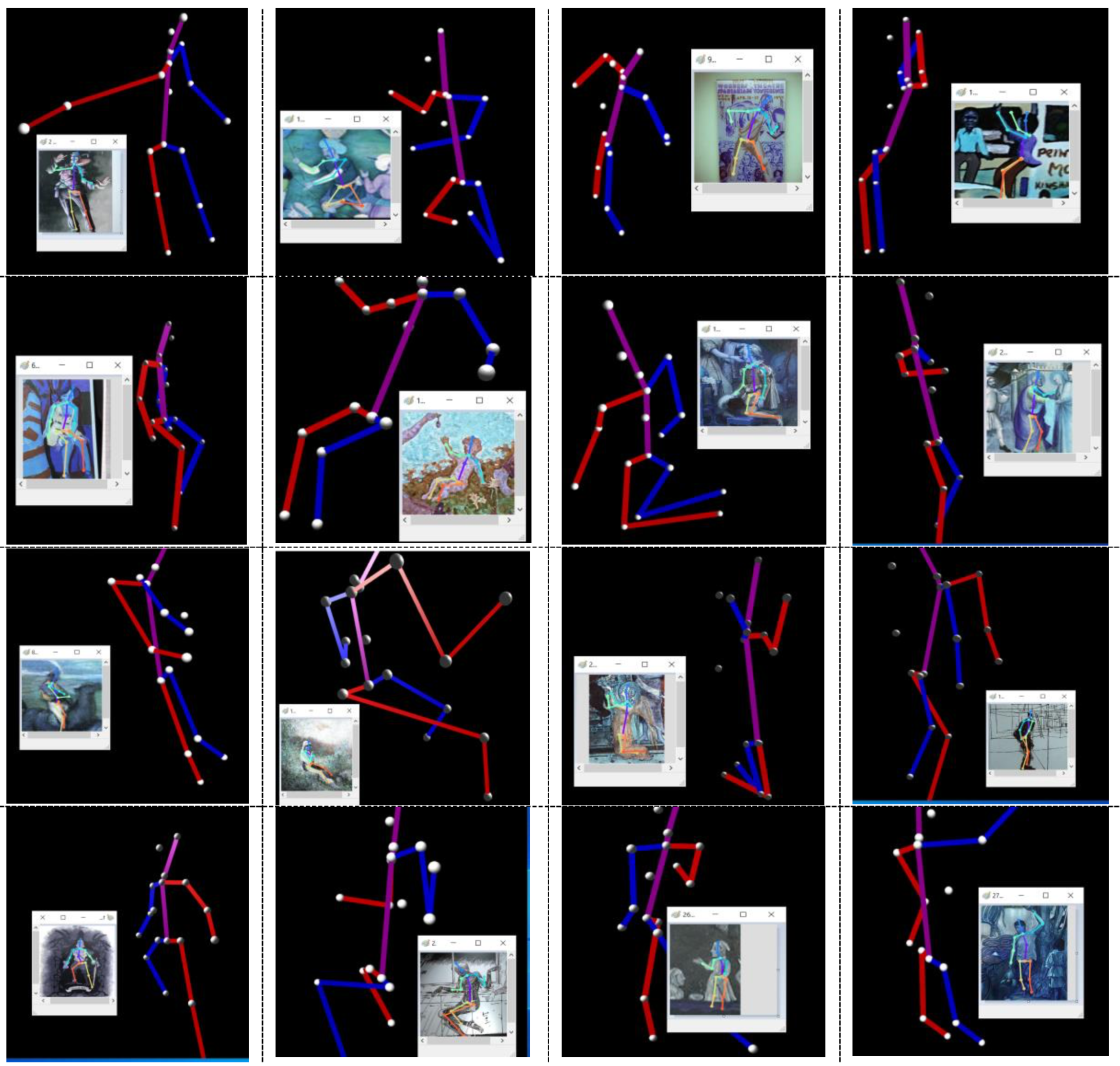}
   
\end{center}
   \caption{2d and 3d predictions on the test set.}
   \label{fig:posepred3d}
\end{figure*}

We show 2d keypoints prediction in Fig. ~\ref{fig:posepred2d}. We simultaneously present regressed 2d and 3d poses in Fig. ~\ref{fig:posepred3d}, where 2d poses are plotted in the small windows. We can see that the 2d and 3d predictions are reasonable.

%
   
\subsection{State-of-the-art Comparison}
\label{sec:sotacomp}
\begin{figure*}
\begin{center}
   \includegraphics[width=0.72\linewidth]{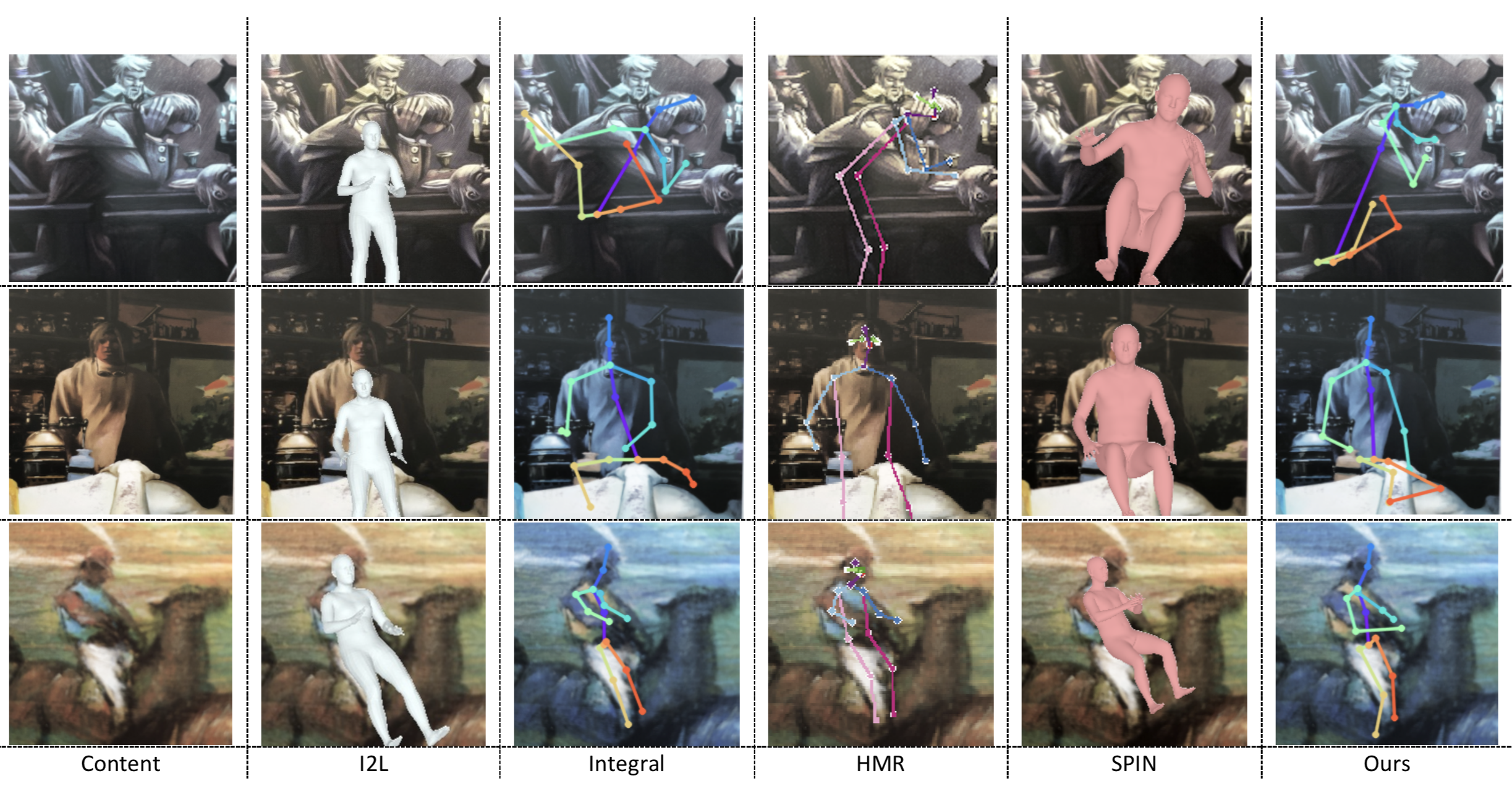}
   
\end{center}
   \caption{Comparison with state-of-the-art methods. \textbf{I2L}: \cite{moon2020i2l}, \textbf{Integral}: \cite{sun2018integral}, \textbf{HMR}: \cite{kanazawa2018end}, \textbf{SPIN}: \cite{kolotouros2019learning}. \textbf{Ours}: our full pipeline.}
   \label{fig:compwsota}
\end{figure*}

 There is no existing work tackling estimation of 3d poses in art images that include a human figure.
In this part, we apply top-performing methods trained on real-world datasets to our test set. 
Results are visualized in Fig. ~\ref{fig:compwsota}. One can easily spot misplaced joints of these methods, which may be attributed to different image domains, lack of diverse pose labels  (\textcolor{magenta}{Appendix Sec. E}).

\subsection{Ablation}

\subsubsection{End-to-end Training vs Per-style Training}

\label{sec:e2etrain}
Per-style training produces more pleasing stylization images than joint-training. In Fig. ~\ref{fig:stylized}, the left $3$ columns correspond to this style-specific training. However, in terms of pose accuracy, end-to-end performs better. To reflect this, we evaluate PCKh@0.25 and draw Table. ~\ref{tab:perstyletrain}.





 


\begin{table*}
\large
\centering
\label{tab:perstyletrain}
\begin{tabular}{l|c|c|c|c|c|c|c|}

\toprule

\textbf{Joint} & {w/o style} & {\emph{la\_muse}} & {3 styles} & {5 styles} & {9 styles} & {12 styles} &  {ours}\\
\hline\
Ankle & 21.4 & 21.5  & 24.7 &  23.8 & 24.6 & 26.0 & \textbf{31.0} \\ \hline
Knee & 23.0 & \textbf{33.5}  & 32.4 &  31.3 & 32.0 & 31.0 & 25.3 \\ \hline
Wrist & 30.9 & 30.1  & 31.7 &  32.4 & 32.4 & 33.4 & \textbf{37.7}   \\ \hline
Elbow & 33.1 & 33.7  & 34.0 &  35.7 & 35.2 & 35.9 & \textbf{44.4}  \\ \hline
Shoulder & 41.0 & 43.0 & 43.0 & 43.7 & 42.5 & 41.2 & \textbf{55.8}   \\ \hline
Head & 52.9 & 51.2  & 55.1 & 55.3 &  55.6 & 55.6 & \textbf{58.9}  \\ \hline
Hip & 19.4 & 21.7 & 21.2 &  22.1 & 23.0 & 23.5 & \textbf{25.8} \\ \hline
Total & 34.5 & 35.3  & 36.9 &  37.4 & 37.6 & 37.9 & \textbf{44.4} \\ 
\bottomrule
\end{tabular}

\caption{PCKh@0.25 (\%) (\textcolor{blue}{the higher, the better}) of \emph{style-specific training} vs \emph{arbitrary style joint training}. \textbf{\emph{la\_muse}}: test the model trained with only style \emph{la\_muse}.
 \textbf{$N$ styles} : average the predictions of $N$ style-specific models on the test set. \textbf{ours}: the full end-to-end pipeline. }

\end{table*}






\subsubsection{Style Transfer Losses}

\label{sec:losses}









\begin{figure}
\begin{center}
   \includegraphics[width=0.75\linewidth]{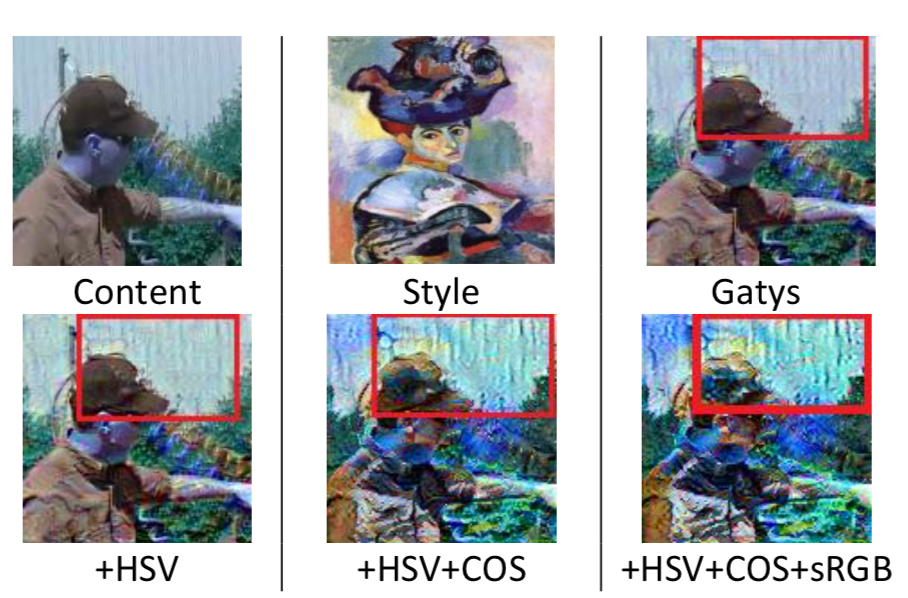}
   
\end{center}
   \caption{Stylized image using different style losses. Adding more improves the production quality. Gatys: \cite{gatys2016image}.}
   
   \label{fig:figlosses}
\end{figure}

Continuing the investigation on style transfer, we now move on to the extra losses introduced. As seen in Fig. ~\ref{fig:figlosses}, adding HSV loss gears the stylized image towards a warm tone,  consistent with the style image. The cosine loss encourages the stylized image to be perceptually similar to the style image in high-level layers but does not necessarily require an exact match. Comparing images in another color space sRGB further boosts the stylized image quality. 

\subsubsection{Pseudo 3d Ground Truth}
\label{sec:pseudo}
\begin{figure}
\begin{center}
   \includegraphics[width=0.85\linewidth]{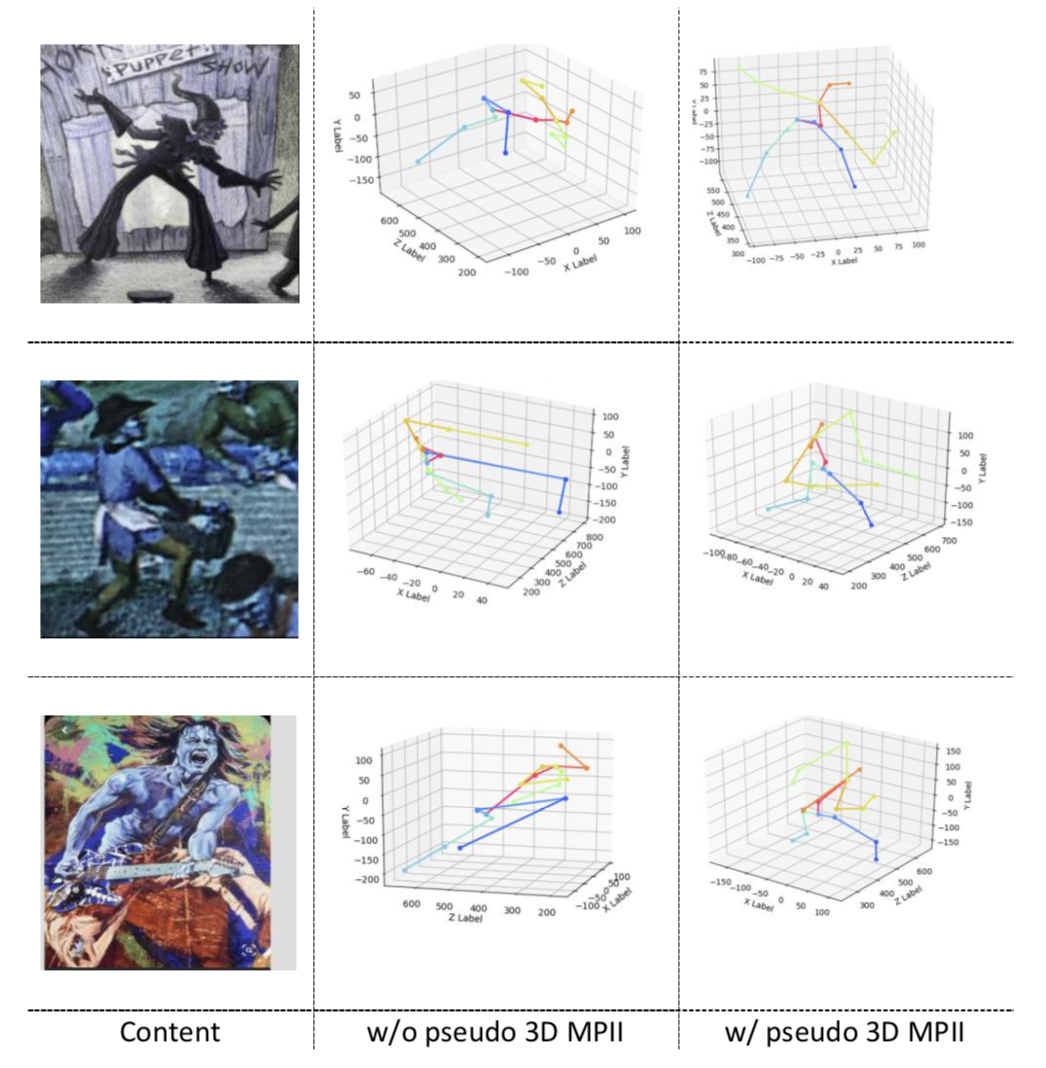}
   
\end{center}
   \caption{Visualization of 3d pose predictions with and without the pseudo 3d ground truth of MPII.}
   
   \label{fig:figmympii}
\end{figure}

 The pose regression part starts to analyze the pose after the stylization is done. As pointed out, to extend the 3d pose variety  (\textcolor{magenta}{Appendix Sec. E}) we manually annotate the pseudo 3d labels of the full MPII. To signify its importance, we compare the predictions trained with and without these labels in Fig. ~\ref{fig:figmympii}. In \emph{row 1},  the left arm is closer to the camera than the right one, which is correctly captured by the model trained with pseudo 3d labels of MPII. In \emph{row 2}, the model preserves the lunge pose with the aid of pseudo 3d labels. Further, in \emph{row 3} the arm prediction without using pseudo 3d labels is arbitrary, whereas the one with pseudo 3d labels is only incognizant of the \emph{forward-leaning} left arm.


\subsubsection{Bone Map}
\label{sec:ske}

\begin{figure}
\begin{center}
   \includegraphics[width=0.85\linewidth]{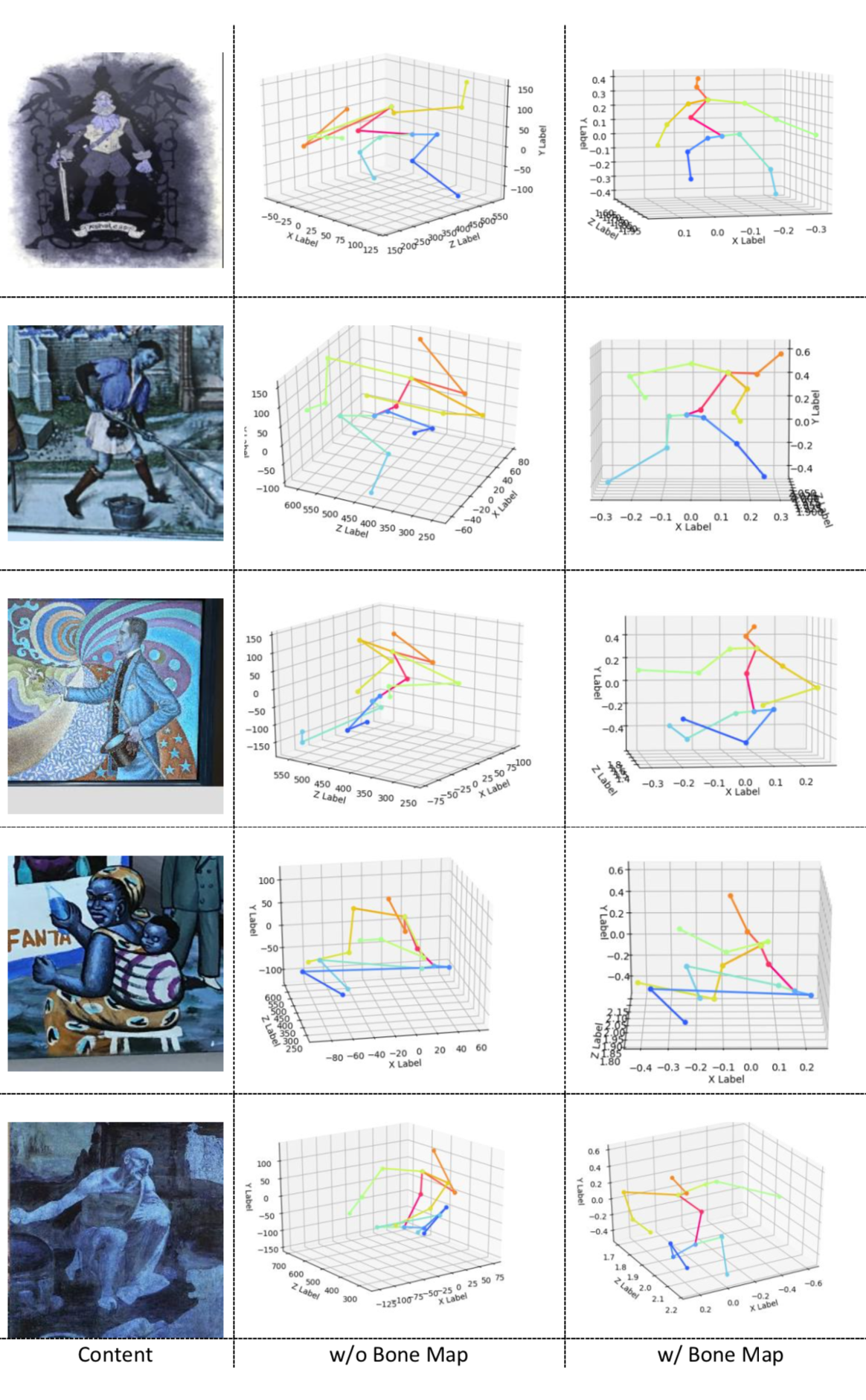}
   
\end{center}
   \caption{3d results with and without bone map.}
   
   \label{fig:figske}
\end{figure}

Extending the discussion on pose regression, here we try to answer how the proposed bone map benefits the depth estimation. Fig. ~\ref{fig:figske} exhibits the pose with and without bone map. It is not hard to observe the mistakes are fixed with the usage of bone map.

\subsubsection{Self Supervision}
The self supervision explicitly improves the 2d pose, which we measure by comparing the PCKh@0.25 of the model with and without self supervision. Looking at Table ~\ref{tab:selfsup}, it is most effective for improving upper body joints $i.e.$ elbow, wrist, shoulder and head, which is understandable since upper body joints are generally easier to estimate. Be aware of that PCKh@0.25 is a stricter threshold than the commonly used PCKh@0.5, but the following is true: self supervision leads to non-trivial improvements. 
\textcolor{magenta}{Appendix Sec. I.2} reflects the property of the bone map space $\mathcal{P}$ and  expresses the bone map from fundamental groups.

\begin{table}
\normalsize
\centering
\begin{tabular}{l|c|c}

\toprule

\textbf{Joint} & {w/o self\_sup} & {w/ self\_sup}\\
\hline\
Ankle & 28.9 & \textbf{29.4} \\ \hline
Knee & 30.7 & \textbf{31.1} \\ \hline
Wrist & 34.7 & \textbf{38.7}  \\ \hline
Elbow & 41.3 & \textbf{44.4}  \\ \hline
Shoulder & 48.1 & \textbf{49.5}  \\ \hline
Head & 58.3 & \textbf{60.2}  \\ \hline
Hip & 27.0 & \textbf{27.3} \\ \hline
Total & 41.7 & \textbf{43.0} \\ 
\bottomrule
\end{tabular}

\caption{PCKh@0.25 (\%) between the model trained with and without self supervision. \textcolor{blue}{The higher, the better.}}
\label{tab:selfsup}
\end{table}






\subsection{Cross-dataset Generalization}
\label{sec:crossdataset}
In this section we will take an excursion into generalizing to another domain: real-world images. The main focus of the project is to evaluate on artistic style human images, yet we would also like to see how it will adapt to real human figures. For this purpose, we apply the model trained exclusively on stylized images to H36M. The MPJPE (w/ Procrustes Alignment) is 82.91 $mm$. It is certainly much worse than state-of-the-art works, but do note that H36M dataset is not used anywhere for this setting. Recall in Sec. ~\ref{sec:sotacomp} we apply methods trained on real images to our artistic test set. One interesting question to ask then would be, \emph{what would happen if we include both real images and artistic images in training?} We adjust the ratio between real and art images in training pose network $\mathcal{G}$, which means a random portion of $\mathbf{O}$ will be replaced with the original RGB $\mathbf{I}$. We load weights of the main system and fine tune from there. (\textcolor{magenta}{Appendix Sec. H}) In Table. ~\ref{tab:ratio}, we see naturally using more real images leads to better accuracy on H36M. Yet we find the result on our test set is only slightly improved. Since the depth is estimated from the bone map by decoupling, we do not see a significant boost. However, having said that, including images from both real and artistic domain would be conducive if we demand a generic pose model that works on any images containing a human. (See \textcolor{magenta}{Appendix Sec. G})

\begin{table}
\small
\centering
\label{tab:ratio}
\begin{tabular}{l|c|c|c|c|}

\toprule

\textbf{Real:stylized ratio} & {H36M PA MPJPE} & {PCKh@0.25}\\
\hline\
10\% & 91.6 &  41.8\\ \hline 
30\% & 71.8 &  44.5\\ \hline 
50\% & \textbf{63.5} & \textbf{45.5} \\ \hline 

70\% & 67.4 & 42.7 \\  
\bottomrule
\end{tabular}

\caption{Quantitative effect on H36M and our test set of different real:artistic  ratios. PA MPJPE (\emph{mm}): \textbf{\emph{the lower the better}}. PCKh@0.25 (\%):  \textbf{\emph{the higher the better.}}}

\end{table}


   
   


\section{Conclusion}
We present the first method to estimate 3d human poses in artworks. We show considerably well results in varied cases. Key to our approach is the customization of neural style transfer and the exploration of pose regression. We exploit neural style transfer to (a) stylize real images for pose regression (b) self supervise 2d pose predictions. We create a solid baseline using per-style training with instance normalization and exhibit notable stylized image collections. Notice this already gives strong pose results. For further improvement, we build an end-to-end system featuring novel style losses, self supervision, bone map. We infuse the system with a self-collected style set and self-annotated pseudo 3d labels of MPII. We build a test set for evaluation and ablation. Along the way, we elevate the system to a generic pose model with real/art balancing. Theory comes up next.

{\small
\bibliographystyle{ieee_fullname}
\bibliography{egbib}

\begin{thebibliography}{10}\itemsep=-1pt

\bibitem{akhter2015pose}
Ijaz Akhter and Michael~J Black.
\newblock Pose-conditioned joint angle limits for 3d human pose reconstruction.
\newblock In {\em Proceedings of the IEEE conference on computer vision and
  pattern recognition}, pages 1446--1455, 2015.

\bibitem{andriluka20142d}
Mykhaylo Andriluka, Leonid Pishchulin, Peter Gehler, and Bernt Schiele.
\newblock 2d human pose estimation: New benchmark and state of the art
  analysis.
\newblock In {\em Proceedings of the IEEE Conference on computer Vision and
  Pattern Recognition}, pages 3686--3693, 2014.

\bibitem{bregler1998tracking}
Christoph Bregler and Jitendra Malik.
\newblock Tracking people with twists and exponential maps.
\newblock In {\em Proceedings. 1998 IEEE Computer Society Conference on
  Computer Vision and Pattern Recognition (Cat. No. 98CB36231)}, pages 8--15.
  IEEE, 1998.

\bibitem{chen2020monocular}
Yucheng Chen, Yingli Tian, and Mingyi He.
\newblock Monocular human pose estimation: A survey of deep learning-based
  methods.
\newblock {\em Computer Vision and Image Understanding}, 192:102897, 2020.

\bibitem{felzenszwalb2008discriminatively}
Pedro Felzenszwalb, David McAllester, and Deva Ramanan.
\newblock A discriminatively trained, multiscale, deformable part model.
\newblock In {\em 2008 IEEE conference on computer vision and pattern
  recognition}, pages 1--8. IEEE, 2008.

\bibitem{fry1978cubism}
E.F. Fry.
\newblock {\em Cubism}.
\newblock World of art. Oxford University Press, 1978.

\bibitem{gatys2016image}
Leon~A Gatys, Alexander~S Ecker, and Matthias Bethge.
\newblock Image style transfer using convolutional neural networks.
\newblock In {\em Proceedings of the IEEE conference on computer vision and
  pattern recognition}, pages 2414--2423, 2016.

\bibitem{DBLP:journals/corr/abs-1902-00267}
Shreyank~N. Gowda and Chun Yuan.
\newblock Colornet: Investigating the importance of color spaces for image
  classification.
\newblock {\em CoRR}, abs/1902.00267, 2019.

\bibitem{habibie2019wild}
Ikhsanul Habibie, Weipeng Xu, Dushyant Mehta, Gerard Pons-Moll, and Christian
  Theobalt.
\newblock In the wild human pose estimation using explicit 2d features and
  intermediate 3d representations.
\newblock In {\em Proceedings of the IEEE Conference on Computer Vision and
  Pattern Recognition}, pages 10905--10914, 2019.

\bibitem{h36m_pami}
Catalin Ionescu, Dragos Papava, Vlad Olaru, and Cristian Sminchisescu.
\newblock Human3.6m: Large scale datasets and predictive methods for 3d human
  sensing in natural environments.
\newblock {\em IEEE Transactions on Pattern Analysis and Machine Intelligence},
  36(7):1325--1339, jul 2014.

\bibitem{jenicek2019linking}
Tomas Jenicek and Ond{\v{r}}ej Chum.
\newblock Linking art through human poses.
\newblock In {\em 2019 International Conference on Document Analysis and
  Recognition (ICDAR)}, pages 1338--1345. IEEE, 2019.

\bibitem{johnson2016perceptual}
Justin Johnson, Alexandre Alahi, and Li Fei-Fei.
\newblock Perceptual losses for real-time style transfer and super-resolution.
\newblock In {\em European conference on computer vision}, pages 694--711.
  Springer, 2016.

\bibitem{kanazawa2018end}
Angjoo Kanazawa, Michael~J Black, David~W Jacobs, and Jitendra Malik.
\newblock End-to-end recovery of human shape and pose.
\newblock In {\em Proceedings of the IEEE Conference on Computer Vision and
  Pattern Recognition}, pages 7122--7131, 2018.

\bibitem{kolotouros2019learning}
Nikos Kolotouros, Georgios Pavlakos, Michael~J Black, and Kostas Daniilidis.
\newblock Learning to reconstruct 3d human pose and shape via model-fitting in
  the loop.
\newblock In {\em Proceedings of the IEEE International Conference on Computer
  Vision}, pages 2252--2261, 2019.

\bibitem{kundu2020self}
Jogendra~Nath Kundu, Siddharth Seth, Varun Jampani, Mugalodi Rakesh,
  R~Venkatesh Babu, and Anirban Chakraborty.
\newblock Self-supervised 3d human pose estimation via part guided novel image
  synthesis.
\newblock In {\em Proceedings of the IEEE/CVF Conference on Computer Vision and
  Pattern Recognition}, pages 6152--6162, 2020.

\bibitem{li2020cascaded}
Shichao Li, Lei Ke, Kevin Pratama, Yu-Wing Tai, Chi-Keung Tang, and Kwang-Ting
  Cheng.
\newblock Cascaded deep monocular 3d human pose estimation with evolutionary
  training data.
\newblock In {\em Proceedings of the IEEE/CVF Conference on Computer Vision and
  Pattern Recognition}, pages 6173--6183, 2020.

\bibitem{lin2014microsoft}
Tsung-Yi Lin, Michael Maire, Serge Belongie, James Hays, Pietro Perona, Deva
  Ramanan, Piotr Doll{\'a}r, and C~Lawrence Zitnick.
\newblock Microsoft coco: Common objects in context.
\newblock In {\em European conference on computer vision}, pages 740--755.
  Springer, 2014.

\bibitem{luo2018orinet}
Chenxu Luo, Xiao Chu, and Alan Yuille.
\newblock Orinet: A fully convolutional network for 3d human pose estimation.
\newblock {\em arXiv preprint arXiv:1811.04989}, 2018.

\bibitem{madhu2020understanding}
Prathmesh Madhu, Tilman Marquart, Ronak Kosti, Peter Bell, Andreas Maier, and
  Vincent Christlein.
\newblock Understanding compositional structures in art historical images using
  pose and gaze priors.
\newblock {\em arXiv preprint arXiv:2009.03807}, 2020.

\bibitem{madhu2020enhancing}
Prathmesh Madhu, Angel Villar-Corrales, Ronak Kosti, Torsten Bendschus, Corinna
  Reinhardt, Peter Bell, Andreas Maier, and Vincent Christlein.
\newblock Enhancing human pose estimation in ancient vase paintings via
  perceptually-grounded style transfer learning, 2020.

\bibitem{moon2020i2l}
Gyeongsik Moon and Kyoung~Mu Lee.
\newblock I2l-meshnet: Image-to-lixel prediction network for accurate 3d human
  pose and mesh estimation from a single rgb image.
\newblock {\em arXiv preprint arXiv:2008.03713}, 2020.

\bibitem{mori2002estimating}
Greg Mori and Jitendra Malik.
\newblock Estimating human body configurations using shape context matching.
\newblock In {\em European conference on computer vision}, pages 666--680.
  Springer, 2002.

\bibitem{omran2018neural}
Mohamed Omran, Christoph Lassner, Gerard Pons-Moll, Peter Gehler, and Bernt
  Schiele.
\newblock Neural body fitting: Unifying deep learning and model based human
  pose and shape estimation.
\newblock In {\em 2018 international conference on 3D vision (3DV)}, pages
  484--494. IEEE, 2018.

\bibitem{sun2017compositional}
Xiao Sun, Jiaxiang Shang, Shuang Liang, and Yichen Wei.
\newblock Compositional human pose regression.
\newblock In {\em Proceedings of the IEEE International Conference on Computer
  Vision}, pages 2602--2611, 2017.

\bibitem{sun2018integral}
Xiao Sun, Bin Xiao, Fangyin Wei, Shuang Liang, and Yichen Wei.
\newblock Integral human pose regression.
\newblock In {\em Proceedings of the European Conference on Computer Vision
  (ECCV)}, pages 529--545, 2018.

\bibitem{ulyanov2016norm}
Dmitry Ulyanov, Andrea Vedaldi, and Victor Lempitsky.
\newblock Instance normalization: The missing ingredient for fast stylization.
\newblock {\em arXiv preprint arXiv:1607.08022}, 2016.

\bibitem{wan2019patch}
Qingfu Wan, Weichao Qiu, and Alan~L Yuille.
\newblock Patch-based 3d human pose refinement.
\newblock In {\em CVPRW}, 2019.

\bibitem{wan2017deepskeleton}
Qingfu Wan, Wei Zhang, and Xiangyang Xue.
\newblock Deepskeleton: Skeleton map for 3d human pose regression.
\newblock {\em arXiv preprint arXiv:1711.10796}, 2017.

\bibitem{wren1997pfinder}
Christopher~Richard Wren, Ali Azarbayejani, Trevor Darrell, and Alex~Paul
  Pentland.
\newblock Pfinder: Real-time tracking of the human body.
\newblock {\em IEEE Transactions on pattern analysis and machine intelligence},
  19(7):780--785, 1997.

\bibitem{zhou2016deep}
Xingyi Zhou, Xiao Sun, Wei Zhang, Shuang Liang, and Yichen Wei.
\newblock Deep kinematic pose regression.
\newblock In {\em European Conference on Computer Vision}, pages 186--201.
  Springer, 2016.

\end{thebibliography}
}

\end{document}


\title{Appendix} 

\maketitle

\begin{appendices}

    \section{Notations}
    \label{sec:note}
    Just to clarify the notations used in the main paper, in Table ~\ref{tab:nottable} we list them. We define $\mathbf{I}$ as the original RGB image. ($\mathbf{I} \in \mathcal{I}$, $\mathcal{I}$ is the space that contains $\mathbf{I}$). $\mathbf{S} \in \mathcal{S}$ as the style image. $\mathbf{O} \in \mathcal{O}$ as the stylized output of the image transform network. During training, the $\mathbf{O}$ is forwarded to the pose network to generate a 3D pose prediction output $\mathbf{Y} \in \mathcal{Y}$. (the corresponding pose label is referred to as $\mathbf{\hat{Y}}$). For the inference phase, $\mathbf{I}$ is passed to generate $\mathbf{Y}$. More generally, the pose network $\mathbf{G}: \mathcal{X} \rightarrow \mathcal{Y}$. ($\mathcal{X} = \mathcal{I} \cup \mathcal{O}$).

\begin{table}[htbp]
\centering
\begin{tabular}{ll}

\toprule

\textbf{Symbol} & \textbf{Meaning} \\
\hline\

$\mathbf{I}$ & the original RGB\\

$\mathbf{S}$ & the style target \\

$\mathbf{O}$ & the style transferred image \\

$\mathbf{O'}$ & the reconstruction of $\mathbf{O}$ via self supervision part  \\

$\mathbf{Y}$ & the predicted 3d pose\\

$\mathbf{\hat{Y}}$ & the ground truth 3d pose\\

$\mathbf{Y_{2d}}$ & the predicted 2d pose \\

$\mathbf{\hat{Y_{2d}}}$ & the ground truth 2d pose \\

$\mathbf{P}$ & the rendered kinematic bone map from $\mathbf{Y_{2d}}$ \\

$\mathbf{Y_{depth}}$ & the predicted joint depth \\

$\mathcal{I}$ & the space where $\mathbf{I}$ sits inside\\

$\mathcal{S}$ & the space that contains $\mathbf{S}$ \\

$\mathcal{O}$ & the space of $\mathbf{O}$ \\

$\mathcal{X}$ & the space of both real and artistic images \\

$\mathcal{Y}$ & the space of poses \\

$\mathcal{P}$ & the space encapsulating $\mathbf{P}$\\

$\mathcal{F}$ & the image transform net to stylize $\mathbf{I}$\\

$\mathcal{G}$ & the pose network that obtains $\mathbf{Y_{2d}}$\\

$\mathcal{R}$ & the parameter-free function to render $\mathbf{P}$ \\

$\mathcal{G'}$ & the pose network to deduce depth from $\mathbf{P}$\\

$\mathcal{F'}$ & the image transform net to reconstruct $\mathbf{O}$\\

\bottomrule
\end{tabular}

\caption{Notation table}
\label{tab:nottable}
\end{table}

   \section{Loss Ratios}

To train such a large network, balancing the losses is of great essence so that not a single one is dominating. We list the ratios in Table. ~\ref{tab:tabratios}.

   \begin{table}[h!]
  \begin{center}
    
    \begin{tabular}{l|l|r}
      \textbf{Loss Type} & \textbf{Loss Name} & \textbf{Ratio}\\
      
      \hline
      \toprule
      
      \multirow{8}{*}{Neural Style Transfer Loss} & $\mathcal{L}_{style}(\mathrm{O},\mathrm{S}) $ & 1.0\\ 
      & $\mathcal{L}_{tv}(\mathrm{O}) $ & 1.0\\ 
   & $\mathcal{L}_{sent}(\mathrm{O},\mathrm{S}) $ & $10^9$ \\
   
   & $\mathcal{L}_{srgb}(\mathrm{O_{srgb}}, \mathrm{S_{srgb}}) $ & $ 200 $ \\
   
   & $\mathcal{L}_{hsv}(\mathrm{O}, \mathrm{S}) $ & $ 300 $ \\
   
   & $\mathcal{L}_{cos}(\mathrm{O}, \mathrm{S}) $ & $ 1000 $ \\
   & $\mathcal{L}_{content}(\mathrm{O}, \mathrm{I}) $ & $ 1.0 $ \\
   & $\mathcal{L}_{cent}(\mathrm{O}, \mathrm{I}) $ & $ 10^6 $ \\

      \bottomrule
      \toprule
      
      \multirow{2}{*}{Pose Regression Loss} & $\mathcal{L}_{2d} $ & 1.0\\ 
      & $\mathcal{L}_{depth} $ & 1000\\ 
      
      \bottomrule
      \toprule
      
      \multirow{3}{*}{Self Supervision Loss} & $\mathcal{L}_{style\_sup}(\mathrm{O'},\mathrm{O}) $ & 0.0035\\ 
      
      & $\mathcal{L}_{cos\_sup}(\mathrm{O'},\mathrm{O}) $ & 0.0035\\ 
      
      & $\mathcal{L}_{feat\_sup}(\mathrm{O'},\mathrm{O}) $ & 0.0035\\ 
      \bottomrule
      
    \end{tabular}
  \end{center}
  \caption{Loss ratio table.}
   \label{tab:tabratios}

\end{table}

   \section{Styles}

\begin{figure}
\begin{center}
   \includegraphics[width=1.0\linewidth]{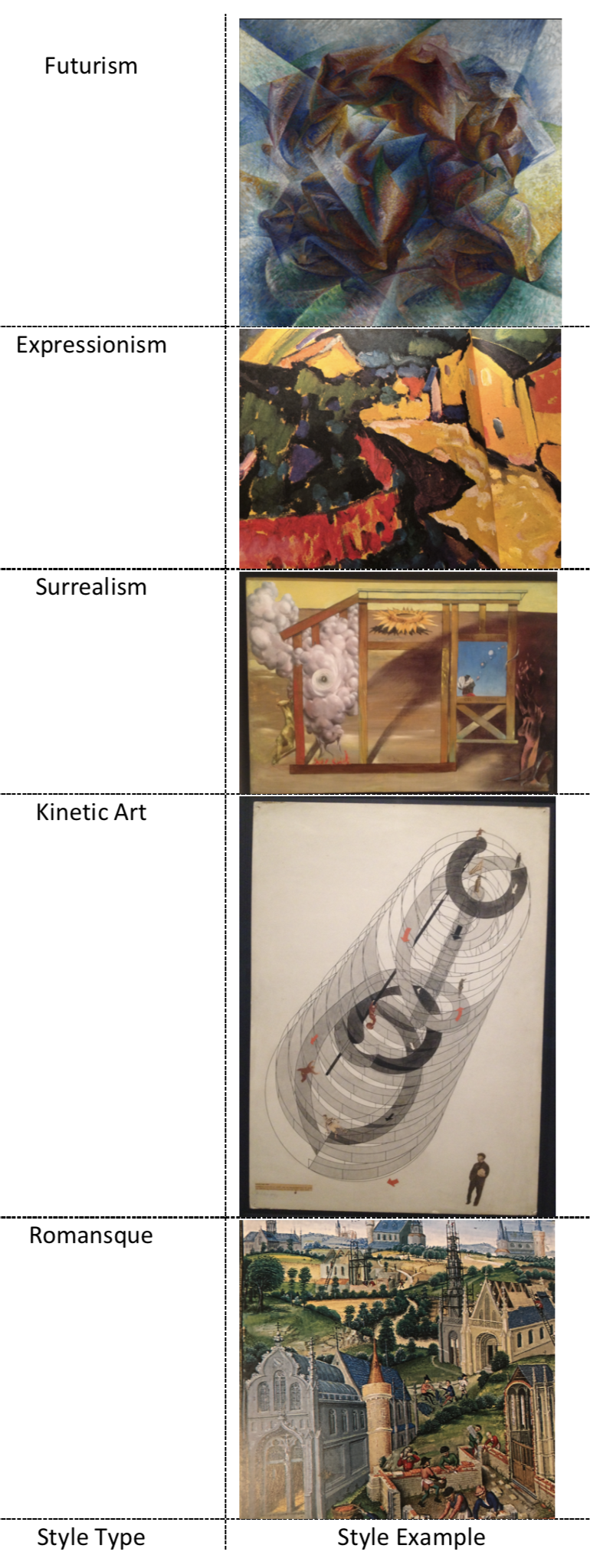}
   
\end{center}
   \caption{Style examples.}
   \label{fig:styles}
\end{figure}

We show some sample images for different styles in Fig. ~\ref{fig:styles}. Please refer to Google Drive for a full view.

    \section{Preliminaries}
	\label{sec:pre}    
	Before analyzing the test cases, we borrow some preliminaries from topology.
	
    \begin{theorem}
    A topological space is an ordered pair ($\mathcal{X}, \mathcal{T}_x$) where $\mathcal{X}$ is a set (of points), $\mathcal{T}_x$ is a collection (family) of subsets satisfying the three following:
    
    \begin{enumerate}
    \item $\emptyset$, $\mathcal{X}$ in $\mathcal{T}_x$ are open.
    
    \item Any (infinite) union of set in $\mathcal{T}_x$ is in $\mathcal{T}_x$: If $\lbrace U_{\alpha} | \alpha \in \mathcal{T}_x \rbrace$ are open sets, then $\cup_{\alpha} U_{\alpha}$ is open.
    
    \item Any intersection of $2$ sets (finite intersection) of open sets are open: If $U$, $V$ are open in $\mathcal{T}_x$, then $U \cap V$ is open.
    \end{enumerate} $\mathcal{T}_x$ is the topology on $\mathcal{X}$, and with $\mathcal{X}$ form the topological space. Usually $\mathcal{T}_x$ is dropped and we will refer to $\mathcal{X}$ as the topological space.
    \end{theorem}
    
    \begin{theorem}
    $\mathbf{G}: \mathcal{X} \rightarrow \mathcal{Y}$ is continuous if and only if $\mathbf{G}^{-1}$(any open subset in $\mathcal{Y}$) is also open in $\mathcal{X}$.
    \end{theorem}
    
	 \begin{theorem}
	The relative topology is defined as follows: if $\mathcal{X}$ has a topological space, and $\mathbf{A} \subset \mathcal{X}$. $\mathcal{X}$ has the topology $\mathcal{T}_{x}$ (a collection of open sets), then $\mathbf{A}$ has the relative topology $\mathcal{T}_A = \lbrace \mathbf{A} \cap \mathbf{U} | \mathbf{U} \in \mathcal{T}_x \rbrace$ 
\end{theorem}	     
   
   \begin{theorem}
   A subset $\mathbf{A} \subset \mathcal{X}$ of a topological space is called compact if whenever $\mathbf{A} \subset \cup_{\alpha} U_{\alpha}$ with $U_{\alpha}$ open in $\mathcal{X}$, then $\mathbf{A} \subset \cup_{j=1}^N U_j$ (finite subcover property).
   \end{theorem}
    
	\begin{theorem}
    A path $\alpha:[0,1] \rightarrow Y$ is a time-parameterized mapping (a function; a "parameterized curve") between the interval $I=[0,1]$ (note this $I$ is different from the original RGB $\mathbf{I}$) and the space $Y$.
    
    \end{theorem}    
    
    \begin{theorem}
    A path $\alpha:[0,1] \rightarrow Y$ is said to be a loop if $\alpha(0) = \alpha(1)$
    
    \end{theorem}
    
    \begin{theorem}
    A path $\alpha:[0,1] \rightarrow Y$ is homotopic to $\beta: [0,1]\rightarrow Y$ written as $\alpha \underset{h}{\leadsto} \beta$ if they have the same end points: $\alpha(0)=\beta(0)$ and $\alpha(1)=\beta(1)$
    
    \end{theorem}

    \section{Test Case Analysis 1: Generalize to Artistic Style Images}

When we say $\mathit{Img_{1}}$,  $\mathit{Img_{2}}$ are similar in a space $\mathcal{X}$ within metric space ($\mathcal{X}$, $d$) for some unknown metric $d$, it is formalized as $d(\mathit{Img_{1}}, \mathit{Img_{2}})< r$ for some $r$ (predefined by as the accetable range radius). This leads us to the {open ball} concept in metric space $B_r(\mathit{Img_{1}}) = \lbrace \mathit{Img_{*}} \in \mathcal{X} | d(\mathit{Img_1}, \mathit{Img_{*}}) < r \rbrace$ is the {open ball} of radius $r$ about $\mathit{Img_1}$ in $\mathcal{X}$ for some unknown distance metric $d$ (\emph{euclidean}, \emph{cosine similarity}, \emph{maximum mean discrepancy} \etc ) All the images (as individual points inside $\mathcal{X}$ in $B_r(\mathit{Img_{1}}) $ is $\mathit{Img_{1}}$'s neighborhood. Let an image $\mathit{Img_1}$ with its corresponding pose output $\mathbf{Y_1}$, and another image $\mathit{Img_2} : \mathit{Img_1} + \mathit{\Delta_1}$ with its pose output $\mathbf{Y_2}$. If $\Delta_1$ is relatively small, namely $\mathit{Img_2}$ is $\mathit{Img_1}$'s neighbor $\mathit{Img_2} \in 
B_r(\mathit{Img_{1}}) $, $\mathbf{Y_2}$ would also be close to $\mathbf{Y_1}$: $\mathbf{Y_2} \in B_{\epsilon}(\mathbf{Y_{2}})$.

However, it introduces $2$ unknown variables: $r$ and distance metric $d$.

This can be abstracted in topology as follows:
write $\mathsf{N}(\mathit{Img_1}) \in \mathcal{T}$ ($\mathcal{T}$ is the topology) as its neighborhood. $\mathsf{N}(\mathit{Img_1}) $ is the open set containing image $\mathit{Img_1}$. The description bypasses the need to use $r$ and $d$, which would later be convenient.

Consider the case where the model performs well on the pose prediction. The ground truth 3D pose (unknown) is $\hat{\mathbf{Y}}_1$. Let $\mathbf{A}$ be an open subset in $\mathcal{Y}$ containing $\hat{\mathbf{Y}}_1$: the neighborhood of the \emph{unknown} pose label $\hat{\mathbf{Y}}_1$. $\mathbf{A}$ can be conceptually interpreted as the acceptable (reasonable) range without rigorous examination of the accuracy. The fact pose prediction is good tells us $\mathbf{G}(\mathbf{I_1}) \subset \mathbf{A}$, $\mathbf{I_1} \subset \mathbf{G}^{-1}(\mathbf{A})$. $\mathbf{G}^{-1}(\mathbf{A})$ is the open set covering the images that produce reasonable pose predictions within the acceptable open set "range" $\mathbf{A}$. 
$\mathbf{G}^{-1}(\mathbf{A})$ is open in $\mathcal{X}$ be definition of $\mathbf{G}: \mathcal{X} \rightarrow \mathcal{Y}$ being continuous. Intuitively this means $\mathbf{I_1}$ is of some distance $\epsilon > 0$ ($\epsilon$ is a small number) to a point in $\mathbf{G}^{-1}(\mathbf{A})$. \emph{(Note we have not yet formally defined what the distance is, but it does not affect the illustration)}.

We write the following $2$ conditions:

\begin{enumerate}
\item[(a)] The pose of $\mathbf{I_1}$ is of some distance $\epsilon_{pose} > 0$ to a pose in $\mathbf{G}^{-1}(\mathbf{A})$.

\item[(b)] The style of $\mathbf{I_1}$ is of some distance $\epsilon_{style} > 0$ to a style in $\mathbf{G}^{-1}(\mathbf{A})$.
\end{enumerate}

\textbf{Meeting the condition}

When both (a) and (b) are met, it naturally follows the feature of $\mathbf{I_1}$ is of some distance $\epsilon > 0$ to the feature of a point in $\mathbf{G}^{-1}(\mathbf{A})$. Following this, $\mathit{I_1}$ is \textbf{close} to some point in $\mathbf{G}^{-1}(\mathbf{A})$.

To meet (a), the literature generally resorts to the following:

\begin{enumerate}
\item Jointly train 2D outdoor dataset with only 2D annotation and indoor 3D dataset with \textbf{limited pose variety} compared to MPII.

\item Use synthetic images which also don't exhibit the pose variety of 2D dataset \eg MPII.

\item Use IMU in the wild but the poses are limited compared to athletic poses presented in MPII \eg diving, skiing, swimming, surfing \etc.
\end{enumerate}

Models trained using either way fail on some extreme and rare 3D pose scenarios of an in-the-wild 2D image, though they have achieved relatively high overall accuracy on most images. \textcolor{magenta}{Since we annotate the pseudo ground truth of MPII, (a) is not an issue.}

If condition (b) is not satisfied, when we restrict the discussion of \textbf{close} in the space $\mathcal{X}$, it cannot be said that 
$\mathbf{I_1}$ is in the neighborhood of $\mathbf{G}^{-1}(\mathbf{A})$. Extending the discussion range to deeper layers of the conv net, if the feature of $I_1$ is of some $\epsilon > 0$ to the feature of a point in $\mathbf{G}^{-1}(\mathbf{A})$ starting from layer index $\mathtt{l}$, we say $\mathbf{I_1}$ is \textbf{close} to   some point in $\mathbf{G}^{-1}(\mathbf{A})$.

\begin{corollary}
This means the convolution network is capable of completely separate the human pose content from the style (background) \emph{arguably}.
\end{corollary}

\section{Test Case Analysis 2: Generalize to the Original Training Images}

Note the training phase refrains from making use of the original training images. (See Sec. ~\ref{sec:note}). In this scenario, training is done on artistic style images while testing is on the original RGB input of H36M subject S9 and S11, following standard practice.

Recall in Sec. ~\ref{sec:note} $\mathcal{X} = \mathcal{I} \cup \mathcal{O}$. $\mathcal{T}_x$ is the fully topology on set $\mathcal{X}$, $\mathcal{T}_x = \lbrace (u) | u \in \mathcal{X} \rbrace$. 

During training, only stylized image $\mathbf{O}$ is forwarded to the pose component, $\mathbf{O}$ has a relative topology $\mathcal{T}_O$ = $\lbrace$ $ \mathbf{O} \cap \mathbf{U} | \mathbf{U} \in \mathcal{T}_x  $ $\rbrace$  (check Sec. ~\ref{sec:pre})

During inference on the original H36M, $\mathbf{I}$ has a relative topology $\mathcal{T}_I = $ $\lbrace$ $\mathbf{I} \cap \mathbf{U} | \mathbf{U} \in \mathcal{T}_x$ $\rbrace$.

Recall in test case 1, when the model performs well, $\mathbf{G}(\mathbf{I}_1) \subset \mathbf{A}$, $\mathbf{I}_1 \subset \mathbf{G}^{-1}(\mathbf{A})$.

From the perspective of metric space, for point $\mathbf{I}_1 \in \mathcal{I}$, for $\forall \epsilon >0$, $\exists$ finite set of points $\mathbf{O}_1, \mathbf{O}_2, ..., \mathbf{O}_k$ $\in$ $\mathcal{O}$ (since we have finite styles and finite images) $s.t.$ $\mathbf{I}_1$ is of distance $< \epsilon$ of at least $1$ of these $\mathbf{O}_j$ ($j=1..k$) points. Equivalently, $\exists$ finite collection of open balls $B_{\epsilon}$ $(\mathit{\mathbf{O}_j})$ which together cover $\lbrace \mathbf{I}_1 \rbrace$. Then $(\mathcal{X}, d)$ is precompact. Unfortunately, $\mathbb{R}^n$ is not precompact because it is not closed. Given the original image $\mathbf{I}_1$, style $\mathbf{S}_1, \mathbf{S}_2, ..., \mathbf{S}_k, ...$, a sequence $\lbrace \mathbf{O}_1, \mathbf{O}_2, ..., \mathbf{O}_k, ... \rbrace$ does not have a limit in itself.

Through the lens of topology, given a cover of $\lbrace \mathbf{I}_1 \rbrace$ by open sets in $\mathcal{X} = \mathcal{I} \cup \mathcal{O}$: $\lbrace U_j \rbrace_{j\in \mathcal{J}}$, we would like to see finitely many of these $U_j$ (open sets in $\mathcal{X}$) cover $\lbrace \mathbf{I}_1 \rbrace$. Since $\mathbf{I}_1 \subset \mathbf{G}^{-1}(\mathbf{A})$ (remember it delivers a good pose prediction), $\lbrace \mathbf{I}_1 \rbrace \subset \cup_{j=1}^k U_j$ ($k$ is the number of open sets). Because during training, we restrict ourselves to the relative topology $\mathcal{T}_O$ (remember the pose network only takes the stylized $\mathbf{O}$ as input), the cover is actually done in the open sets of $\mathbf{O}$.

The finite subcover property as discussed in the last paragraph gives an important topological property called compactness.

Based on the fact 

\begin{enumerate}
\item $\mathcal{X}, \mathcal{Y}$ are Hausdorff spaces

\item  $\mathcal{I}$  is compact
\end{enumerate}
, $\mathbf{G}(\mathcal{I})$ is also compact, given $\mathbf{G}:\mathcal{X} \rightarrow \mathcal{Y}$ is a $1$-$1$ correspondence, we derive that $\mathbf{G}$ is a homeomorphism.

The indication is that: \textcolor{magenta}{given the original image $\mathbf{I}_1$ and its stylized variants which together cover it (\emph{finite subcover}) $\mathbf{O}_1, \mathbf{O}_2, ... \mathbf{O}_k$ ($k$ styles), as well as the pose label $\hat{\mathbf{Y}}$, it follows then $\mathbf{G}(\mathbf{I}_1)$ is the neighbor to $\mathbf{G}(\mathbf{O}_1)$,  $\mathbf{G}(\mathbf{O}_2)$, ..., $\mathbf{G}(\mathbf{O}_k)$. (\emph{in the same open set})}.

\begin{corollary}
They ($\mathbf{I}_1$ and $\mathbf{O}_1$, ..., $\mathbf{O}_k$) are similar in visual content. Again, as in test case 1, restricting ourselves to the space $\mathcal{X}$ does not guarantee that. Because this would imply the stylized image is too close to the real-world images, which nullifies the stylization process. Contradiction. The only way to get around this is to extend the space where closeness is analyzed to deeper feature layers of the conv network. 

\end{corollary}

\section{Summary on 2 Test Cases}
If the model outputs reasonably well pose predictions on both $\mathbf{I}_1$ and $\mathbf{O}_1$ with the same pose label $\hat{\mathbf{Y}_1}$, it means either of the following holds:

\begin{enumerate}
\item $\mathbf{I}_1$ is too close to $\mathbf{O}_1$ in space $\mathcal{X}$ (recall the corollary part of test case 2). Namely, the style transfer procedure fails completely and degenerates to have the same (or very similar) output as input.

\item The pose network $\mathbf{G}$ \emph{arguably} yields similar features at some layer $\mathtt{l}$ (and of course layers onwards) for $\mathbf{I}_1$ and $\mathbf{O}_1$. If we extend the discussion as did in test case 1 and 2: extend the map $\mathbf{G}$ from $\mathcal{X} \rightarrow \mathcal{Y}$ to $\mathbf{H}: \mathcal{X} \rightarrow \mathcal{V}$ and $\mathbf{M}: \mathcal{V} \rightarrow \mathcal{Y}$. ($\mathbf{G} = \mathbf{M} \circ \mathbf{H}$ for some intermediate feaute space $\mathcal{V}$) Then $(\mathcal{V},d)$ is precompact from metric point of view, and also $\mathbf{M}: \mathcal{V} \rightarrow \mathcal{Y}$ is now a homeomorphism $\Rightarrow$ $U$ open in $\mathbf{V}$ $\Leftrightarrow$ $\mathbf{M}(U)$ open in $\mathbf{Y}$.
\end{enumerate}

\textbf{Meeting the condition} To achieve condition $2$, there are several ways:

\begin{enumerate}
\item Jointly train the original image with the stylized counterpart within a batch, namely each batch is a mixture of both original and stylized ones. \textcolor{magenta}{This is what we did for the cross-dataset generalization experiment in the main body}.

\item Use a siamese network, and pass both $\mathbf{O}$ and its related $\mathbf{I}$ as input.

\item Pass both $\mathbf{I}$ and $\mathbf{O}$ to two separate streams and enforce similarity losses at certain layers.

\item Decouple the problem into 2D plus depth. 2D is first estimated (which contains a certain degree of depth ambiguity), from which depth is deduced.
\end{enumerate}

We subscribe to the fourth solution in the pipeline.
\section{Joint Training of Real-world Images and Stylized Images}

\textbf{Homotopic paths} Use the above notation, $\mathbf{I_1}$ is the original H36M rgb image, $\mathbf{O_1}$ is some stylized image of $\mathbf{I_1}$, while $\mathbf{Y_1}$ is the ground truth pose. If the input to the pose network is the artistic style transferred image, the training goes through the path $\alpha$ in \textbf{Fig.}: $\mathbf{I_1} \rightarrow \mathbf{O_1} \rightarrow \mathbf{Y_1}$. If otherwise, the input is the original RGB image, then training flows through the path $\beta: \mathbf{I_1} \rightarrow \mathbf{Y_1}$. If both paths produce the same final outcome: the pose prediction for the same original image, (\emph{which we expect to see since they correspond to the same image, and thus the same pose label}), we say path $\alpha$ is homotopic to path $\beta$ $i.e.$ $\alpha \underset{h}{\leadsto} \beta$.

The homotopy $\mathbf{H}$ is 

\begin{enumerate}
\item $H(0,s) = \mathbf{I_1} , 0 <s<1$

\item $H(1,s) = \mathbf{Y_1} , 0 <s<1$

\item $H(t_0,1) = \mathbf{O_1}$ $t_0$ is the moment the pose part starts (to take input images), $0<t_0<1$

\item $H(t_0+\Delta_t,1)=$ \emph{features of layer index $\frac{\Delta_t}{1-t_0}*$ total number of layers}, $0 \leq Delta_t < 1-t_0$

\item $H(t,0)=$ \emph{features of layer index $t*$ number of totla layers}, $0<t<1$.
\end{enumerate}

We observe that the path from $\mathbf{O_1}\rightarrow \mathbf{Y_1}$ is moving at a faster speed than $\mathbf{I_1} \rightarrow \mathbf{Y_1}$.

This poses a challenge when jointly training them, since one can imagine in the path $\alpha$ the original image $\mathbf{I_1}$ is first diverted dramatically to a distant image $\mathbf{O_1}$ (\emph{otherwise the style transfer does not take effect}), then the pose component tries to pull it back to the apropos track so that it has the same output as the original image $\mathbf{I_1}$. Issues about gradient magnitude may arise. In the cross-dataset generalization section of the main paper, we finetune the weights trained exclusively on stylized images to initiate the "joint training" of both real and stylized images.

\section{Self Supervision}

\textbf{Motivation}
The motivation follows the analysis-by-synthesis mechanism where the output usually is mapped to the original image space for self supervision. In our scenario, the output (either stylized $\mathbf{O}$ or joint prediction $\mathbf{Y}$ is projected to the input image space $\mathcal{I}$, from which losses can be enforced.)

\textbf{Quotient space} Given space $\mathcal{O}$ (Sec. ~\ref{sec:note}) (stylized) and equivalence relationship $\sim$ among the points in $\mathcal{O}$, the equivalence relationship is defined as $\mathbf{O_1} \sim \mathbf{O_2}$ if $\mathbf{O_1}$ and $\mathbf{O_2}$ are stylized using the same image content $\mathbf{I_1}$. $\left[ \mathbf{O}\right] $ is the equivalence class of $\mathbf{O}$ in the set of equivalence classes of points of $\mathcal{O}$.

\textbf{Path-connectedness} We would like the propagated gradients to flow smoothly through the input space (\emph{wherever the image resides for the self supervision part}). In the following, the first attempt does not satisfy this while the second one meets this requirement.

\subsection{Reconstructing $\mathbf{I}$ from $\mathbf{O}$} 
This effectively means $\mathbf{O} \stackrel{q}{\longrightarrow} \mathbf{O}/\sim$
 $\stackrel{r}{\longrightarrow} \mathbf{I}$: we first map $\mathbf{O}$ to the equivalence class where each image is derived from the same original image content. However, it is practically difficult to map $r$.
 
\textbf{Remark} The space $\mathcal{X}=\mathcal{I} \cup \mathcal{O}$ is not path-connected because path-connectedness implies $\forall a,b\in \mathcal{X}$, $\exists$ a path $\omega: [0,1]\rightarrow \mathcal{X}$ from $a$ to $b$. In this case let $a=\mathbf{O}$ and $b=\mathbf{I}$, it's hard to find a path from $\mathbf{O}\rightarrow \mathbf{I}$.

Measuring the difference in the original image space is difficult due to geometric and photometric transformations. We instead opt to \emph{implicitly optimize the stylization} (via network) again and gauge the discrepancy between stylization $\mathbf{O}$ and re-optimized stylization $\mathbf{O}^\prime$.

\subsection{Reconstructing $\mathbf{O}$ from $\mathbf{S}$ and $\mathbf{Y_{2d}}$} Another approach is to render a 2D bone map $\mathbf{P}$ 
from the 2D pose prediction $\mathbf{Y_{2d}}$, together with the style image $\mathbf{S}$, an image transform network $\mathcal{F}^\prime$ is used to reconstruct the stylized output: $\mathbf{O}^\prime$, which we did in the main algorithm.

Suppose the ground truth pose is $\hat{\mathbf{Y}_{2d}}$. 

\begin{enumerate}
\item If $\mathbf{Y}_{2d}$  is not in $\hat{\mathbf{Y}_{2d}}$'s neighborhood (the open set that contains $\hat{\mathbf{Y}_{2d}}$: $\hat{\mathbf{Y}_{2d}} \in \mathbf{A} \subset \mathcal{Y}_{2d}$; $\mathbf{Y}_{2d} \notin A$.

Then the reconstructed output $\mathbf{O}^\prime$ has the pose content astray from the label $\hat{\mathbf{Y}_{2d}}$. 
 (\emph{The spatial correspondence between $\mathbf{O}^\prime$ and $\mathbf{P}$ is roughly preserved by FCN}) This imposes a large $\mathcal{L}_{cos\_sup}$ which could flow back to $\mathbf{Y}_{2d}$

\item If $\mathbf{Y} \in A$, note the space $\mathcal{P}$ of the rendered 2D bone map has a nice property: it's locally path-connected. That is, for each $p \in \mathcal{P}$, $\exists$ an open set that contains $p$.  Further, it is path-connected, for $\forall p,q \in \mathcal{P}, \exists$ a path $\omega: [0,1]\rightarrow \mathcal{P}$ from $p$ to $q$.

Intuitively this can be explained by the fact we can deduce 2D from such a clean bone map and perform 2D deformation thereafter.

Specifically, consider 2 adjacent bones in the kinematic chain (for instance \emph{pelvis}$\rightarrow$\emph{left\_hip} and \emph{left\_hip}$\rightarrow$ \emph{left\_knee}) connecting $(x_1,y_1)\rightarrow(x_2,y_2)$ and $(x_2,y_2)\rightarrow(x_3,y_3)$. The path from $(x_1,y_1)$ to $(x_2,y_2)$ induces a homeomorphism $f_{*}$ that maps the loop (\emph{the first bone}) $\hat{\alpha_{n}}^1$ (connecting $(x_1,y_1)$ to $(x_2,y_2)$ and back to $(x_1,y_1)$) in the fundamental group $\pi_1(\mathcal{P},(x_1,y_1))$ to the loop (\emph{the second bone}) $\hat{\alpha_{n}}^2$ ($(x_2,y_2) \rightarrow (x_3,y_3) \rightarrow(x_2,y_2)$) in $\pi_1(\mathcal{P},(x_2,y_2))$:

$f_{*}([\hat{\alpha_{n}}^1]) = [\hat{\alpha_{n}}^2]$, which basically means \emph{the first bone} is mapped to \emph{the second bone}. Continue this process, bones further down in the kinematic chain can be mapped by the induced homeomorphism in the fundamental group.

The fundamental group $\pi_1(\mathcal{P},(x_1,y_1))$ is isomorphic ($\cong$) to the fundamental group $\pi_1(\mathcal{P},(x_2,y_2))$. In particular, the fundamental group $\pi_1(\mathcal{P})$ does not depend upon the choice of base point. Which means there is a loop that starts and ends at any pixel inside the bone map (\emph{interior or on edge of the oval areas pertaining to each bone}): the trivial loop $e_x$.

At a high level, the 2D bone map is nothing but a loop based at $x_0$ (since the base point doesn't matter, let's assume for now $x_0$ = \emph{pelvis}) in the fundamental group $\pi_1({\mathcal{P}})$. The idea of continuously deforming the 2D bone map according to backward gradient signal is just:

\begin{itemize}
\item[$\bullet$] Using induced homeomorphism by paths connecting 2 points $(x_1,y_1)$ and $(x_2,y_2)$ to do mapping between loops (\emph{bones}) in the fundamental group.

\item[$\bullet$] Using path-connectedness of space $\mathcal{P}$ and the homeomorphisms to construct the new bone map bone by bone.

\end{itemize}


Then, it follows that 

 \begin{enumerate}
 \item If $\mathbf{Y} = \hat{\mathbf{Y}}$, 2D bone map overlaps entirely with the stylized image $\mathbf{O}$. For the background pixel of the rendered $\mathbf{P}$, the gradient has no impact on the pose prediction $\mathbf{Y}$. For the overlapped part, the style input $\mathbf{S}$ will guide the reconstruction to have the style as $\mathbf{S}$, and so won't incur a large pixel difference loss.
 
 \item If $\mathbf{Y} \neq \hat{\mathbf{Y}}$ but $\mathbf{Y} \in \mathbf{A}$ (in $\hat{\mathbf{Y}}$'s neighborhood), the loss would be small. As described earlier, the style is not an issue, and the background pixel of rendered 2D bone map $\mathbf{P}$ does not affect $\mathbf{Y}$. 
 Since $\mathbf{Y}$ is in $\hat{\mathbf{Y}}$'s neighorbood, $\exists$ a path $\omega$ from $\mathbf{Y}$ to $\hat{\mathbf{Y}}$. For the overlapped pixels, each pixel is in some loop determined by $\hat{\alpha_i}(t)$ for some $t$. And the pixel value is parameterized by the $2$ consecutive keypoints in the kinematic chain. The backpropagation ensures $\mathbf{Y}$ is moving towards $\hat{\mathbf{Y}}$ along the path as iterations increase. (\emph{recall the bone map space $\mathcal{P}$ is path-connected.)}
 \end{enumerate}
\end{enumerate}

We empirically found in our experiments that with the help of neural style transfer, the 2D prediction  $\mathbf{Y_{2d}}$ is already very close to the ground truth $\hat{\mathbf{Y_{2d}}}$, meaning (b) is met. The self supervision loss slightly refines the \textbf{2D output} by virtue of 

\begin{itemize}
\item The \emph{path-connectedness} of space $\mathcal{P}$.

\item The stylization loss. (It does not require an exact per-pixel match between $\mathbf{O}$ and $\mathbf{O}^\prime$)
\end{itemize}

\section{Per-style Training}

The ablation study about per-style training trains parallel independent style transfer networks, upon which we condition independent pose networks to deliver the final pose output. It is possible to use boosting algorithm like Adaboost to fix previous mistakes made by previous styles, meaning the training of later styles depends on earlier styles.

\section{Bone Map}

We continue the definition of bone map here. With regards to the loop, go to \textcolor{magenta}{Preliminaries Definition D.5/D.6}.

For the loop $\hat{\alpha_i}$, the starting point $\hat{\alpha_i}(0)$=the ending point $\hat{\alpha_i}(1)$=$(x_1,y_1)$, the halfway midpoint $\hat{\alpha_i}(\frac{1}{2})=(x_2,y_2)$. For other $t$, the mapped point in space $\mathit{R}^2$: $\hat{\alpha_i}(t)$ is determined such that it satisfies one of the following case:

\begin{enumerate}
\item[$\blacktriangleright$] $0 \leq t \leq \frac{1}{4}$: $t=\frac{dist((x,y),\vec{x})}{4a_i}$

\item[$\blacktriangleright$]
$\frac{1}{4} < t \leq \frac{1}{2}$: $t=\frac{1}{4}+(\frac{1}{4}(1-\frac{dist((x,y),\vec{x})}{a_i}))$

\item[$\blacktriangleright$] $\frac{1}{2} < t \leq \frac{3}{4}$: $t = \frac{1}{2}+\frac{dist((x,y),\vec{x})}{4a_i}$

\item[$\blacktriangleright$] $\frac{3}{4} < t \leq 1$: $t = \frac{3}{4}+(\frac{1}{4}(1-\frac{dist((x,y),\vec{x})}{a_i}))$ 
\end{enumerate} 

\section{Non-exhaustive standalone notes about artistic human figure images}

\begin{itemize}

\item At the beginning of the 14-th century, proportions of human body were lengthened.

\item The 15-th century was a time of change, the rise in realism became noticeable in people's attitude towards the human condition.

\item 
Impressionism concentrates on the landscape rather than on the human figure. People become part of nature.

\item Expressionism uses bolder and brighter colors than impressionism and focuses more on human emotions and expressions. Subject and style are highly simplified. Colors become brighter and more arbitrary, and are arranged to look more powerful.

\item According to virginia woolf, on or about Dec 1910 human nature changed. Scientists had disproved many of the ideas on which nature's laws appeared to have been based and there was a degree of questioning and hypothesizing in all areas of human activities unparalleled in history. The new view of the world had begun to be reflected in the visual arts. Many of the modernist styles had already been created. 
\emph{Fauvism, Cubism, and Expressionism all put forward their own versions of reality. Each in its own way had questioned the tenets of Impressionism.}

\item Cubists like Picasso broke away from two central characteristics of European painting since the Renaissance, one of which being \emph{the classical norm for the human figure}.

\item 
Some painters like Renoir never wandered in the sense of structure and topography, on the other hand, some painters particularly pronounce the shapelessness of their paintings. And so some human figures are more amorphous than others with anatomical constraints completely forgone \eg \emph{Franc Marc}.

\item As time goes, some think abstract art is the only form of modern art while some are in search of pure colors. A new attitude towards the expressive potentialities of the human figure is not on gesture but on complete freedom, meaning unexpressed states of minds are unleashed, which results in surrealism.

\item 
Some paintings are softer/sweeter colors, some are darker/freer/richer.

\end{itemize}

\end{appendices}
